\documentclass[review,authoryear]{elsarticle}

\usepackage{hyperref}
\usepackage{amsmath}
\allowdisplaybreaks
\usepackage{breqn}
\usepackage{calc}  
\usepackage{enumitem}  
\usepackage{amsfonts}
\usepackage{mathrsfs}
\usepackage{xcolor}
\usepackage{natbib}
\usepackage[pagewise]{lineno}
\usepackage{bigstrut}
\usepackage{booktabs}
\usepackage{rotating}
\usepackage{tabularx}
\usepackage{makecell}
\usepackage{hyperref}
\usepackage[ruled,vlined]{algorithm2e} 

\usepackage[top=3.5cm,bottom=3.5cm,left=3.5cm,right=3.5cm]{geometry}
\usepackage{caption}
\usepackage{subcaption}
\usepackage{tikz}
\modulolinenumbers[5]
\usepackage{graphicx, amsfonts,amsmath , boldline, epstopdf,amssymb }
\usepackage{tabu, multirow,multicol ,booktabs ,setspace ,algcompatible ,mathrsfs,dsfont, array, boldline, makecell, booktabs }
\usepackage{url}
\usepackage{epstopdf}
\usepackage{adjustbox, stackengine, color, colortbl, epsfig, cases, enumitem, mathtools, tikz, verbatim}
\usetikzlibrary{shapes,arrows.meta,decorations}
\definecolor{Abi}{rgb}{0.309803, 0.58039, 0.80392}
\definecolor{orange}{rgb}{1, 0.5019, 0}
\definecolor{Red}{rgb}{1, 0, 0}
\definecolor{orange}{rgb}{1, 0.50196, 0}
\definecolor{greenJ}{rgb}{0, 0.6590, 0.42}
\definecolor{Brown}{rgb}{0.588, 0.294, 0}
\definecolor{um}{rgb}{0.0824, 0.1294, 0.4196}
\definecolor{abikam}{rgb}{0.51, 0.93,  0.992}
\usepackage{amsthm}

\usepackage{lineno,hyperref}
\usepackage{amsmath}
\usepackage{breqn}
\usepackage{calc}  
\usepackage{enumitem}  
\usepackage{amsfonts}
\usepackage{mathrsfs}
\usepackage{multirow}
\usepackage{bigstrut}
\usepackage{booktabs}
\usepackage{rotating}
\usepackage{hyperref}
\usepackage{nomencl}

\makenomenclature
\renewcommand\nomgroup[1]{%
  \item[\bfseries
  \ifstrequal{#1}{I}{Sets and Indices}{%
  \ifstrequal{#1}{P}{Parameters}{%
  \ifstrequal{#1}{L}{Functions}{%
  \ifstrequal{#1}{V}{Variables}{}}}}%
]}

\usepackage{bm}
\usepackage[top=3.5cm,bottom=3.5cm,left=3.5cm,right=3.5cm]{geometry}
\usepackage{caption}
 
\modulolinenumbers[5]
\usepackage{mathtools}
\usepackage{stackengine}

\usepackage{lineno,hyperref}
\usepackage{amsmath}
\usepackage{breqn}
\usepackage{calc}  
\usepackage{enumitem}  
\usepackage{amsfonts}
\usepackage{mathrsfs}
\usepackage{multirow}
\usepackage{bigstrut}
\usepackage{booktabs}
\usepackage{rotating}
\usepackage{hyperref}
\usepackage[top=3.5cm,bottom=3.5cm,left=3.5cm,right=3.5cm]{geometry}

\usepackage{xcolor}
\usepackage{soul}
\usepackage{comment}

\usepackage{amsmath, amssymb}

 \usepackage{caption}

\usepackage{color}

\newcommand{\reals}{{I\kern-.35em R}}
\newcommand{\nats}{{I\kern-.35em N}}
\newcommand{\upto}{{\raise 1pt \hbox{$\scriptstyle \,\nearrow\,$}}}
\newcommand{\downto}{{\raise 1pt \hbox{$\scriptstyle \,\searrow\,$}}}
\newcommand{\eop}
	{\hfill{$\vcenter{\hrule height1pt \hbox{\vrule width1pt height5pt 
   	 \kern5pt \vrule width1pt} \hrule height1pt}$} \medskip}
\def\state #1. { \noindent{\bf#1.\enspace}}

\usepackage{lineno,hyperref}
\usepackage{amsmath}
\usepackage{breqn}
\usepackage{calc}  
\usepackage{enumitem}  
\usepackage{amsfonts}
\usepackage{mathrsfs}

\usepackage{xcolor}
\usepackage{bigstrut}
\usepackage{rotating}
\usepackage[top=3.5cm,bottom=3.5cm,left=3.5cm,right=3.5cm]{geometry}

\usepackage{subcaption}
\modulolinenumbers[5]
\usepackage{graphicx, amsfonts,amsmath , boldline, epstopdf,amssymb }
\usepackage{tabu, multirow,multicol ,booktabs ,setspace ,algcompatible ,mathrsfs,dsfont, array, boldline, makecell, booktabs }
\usepackage{url}
\usepackage{epstopdf}
\usepackage{adjustbox, stackengine, color, colortbl, epsfig, cases, enumitem, mathtools, tikz, verbatim}
\usetikzlibrary{shapes,arrows.meta,decorations}
\definecolor{um}{rgb}{0.0824, 0.1294, 0.4196}
\definecolor{Abi}{rgb}{0.059, 0.322, 0.729}
\definecolor{orange}{rgb}{1, 0.347, 0}
\definecolor{c}{rgb}{0, 1, 1}
\definecolor{m}{rgb}{1, 0, 1}

\definecolor{bb}{rgb}{0.2941, 0.5447, 0.7494}
\definecolor{greenJ}{rgb}{0, 0.6590, 0.42}
\definecolor{zereshk}{rgb}{0.588,0,0.098}
\definecolor{Maroon}{rgb}{0.502, 0, 0}
\definecolor{Brown}{rgb}{0.588, 0.294, 0}
\definecolor{Olive}{rgb}{0.502, 0.502, 0}
\definecolor{Navy}{rgb}{0, 0, 0.502}
\definecolor{Orange}{rgb}{1,0.647,0}
\definecolor{Yellow}{rgb}{0.502, 1, 0}
\definecolor{Green}{rgb}{0, 0.502,  0}
\definecolor{Blue}{rgb}{0, 0,0.761}
\definecolor{Lime}{rgb}{0.196, 0.804, 0.196}
\definecolor{Purple}{rgb}{0.502,0,0.502}
\definecolor{Violet}{rgb}{0.561,0,1}
\definecolor{Magneta}{rgb}{1,0,1}
\definecolor{Red}{rgb}{1,0,0}
\definecolor{Gray}{rgb}{0.502, 0.502, 0.502}
\definecolor{abikam}{rgb}{0.51, 0.93,  0.992}
\definecolor{abizeyad}{rgb}{0.16, 0.573,0.761}
\definecolor{rr}{rgb}{0.9047, 0.1918, 0.1988}

\definecolor{orange2}{rgb}{0.85,0.33,0.10}
\definecolor{yellow4}{rgb}{0.93,0.69,0.13}
\definecolor{purple6}{rgb}{0.49,0.18,0.56}
\definecolor{green32}{rgb}{0.47,0.67,0.19}
\definecolor{blue13}{rgb}{0.30,0.75,0.93}
\definecolor{zereshk16}{rgb}{0.64,0.08,0.18}
\definecolor{orange25}{rgb}{0.93,0.69,0.13}
\definecolor{gray27}{rgb}{0.0824, 0.1294, 0.4196}

\usetikzlibrary{tikzmark}

\tikzset{mycircled/.style={circle,draw,inner sep=0.1em,line width=0.04em}}

\tikzset{mycircled2/.style={circle,draw,inner sep=0.1em,line width=0.09em}}

 \usepackage{caption}
 
\journal{Transportation Research Part C: Emerging Technologies}

\begin{document}









\doublespacing
\begin{frontmatter}
\title{A Graph Learning Approach for Network-Wide Real-Time Traffic Flow Prediction During Wildfire Evacuations}



\author{Md Nafees Fuad Rafi\corref{mycorrespondingauthor}}
\author{Samiul Hasan}

\address{Department of Civil, Environmental and Construction Engineering, University of Central Florida, 12800 Pegasus Drive, Orlando, FL 32816, United States}

\cortext[mycorrespondingauthor]{Corresponding author: mdnafeesfuad.rafi@ucf.edu}

\doublespacing

\begin{abstract}
Wildfire evacuations can cause abrupt network-wide traffic conditions, leading to congestion and safety risks for evacuees. Knowing the traffic conditions in advance can improve overall evacuation traffic management. However, limited availability of wildfire evacuation traffic data poses a challenge for training reliable traffic prediction models. It is therefore important to evaluate whether models trained on regular traffic data can generalize to unseen wildfire evacuations. This study develops a graph learning-based model using routinely available traffic data for network-wide real-time traffic flow prediction during wildfire evacuations. The framework integrates Graph Transformer (GT) and Long Short-Term Memory (LSTM) networks into a unified GT--LSTM framework. After constructing a detector network graph, a spatial fusion gate was introduced to combine local and upstream detector features. Additionally, a learnable detector embedding representation was integrated to capture location-specific traffic patterns. The model was trained using regular period traffic data from October--December 2024 and evaluated during the peak evacuation period of the Palisades and Eaton wildfires in Los Angeles on January 7--8, 2025. The model achieved RMSE of 292 and an $R^2$ of 0.98 for next 1-hour prediction. Across the 1- to 6-hour prediction horizons, it achieved an overall RMSE of 435 and $R^2$ of 0.96. The proposed model outperformed all state-of-the-art baseline models. Furthermore, ablation experiments confirmed that each major model component contributed to improving the prediction performance. The framework can help transportation agencies anticipate traffic congestion, identify potential bottlenecks, and implement proactive traffic management strategies to improve evacuation safety and efficiency.
\end{abstract}
\begin{keyword}
Wildfire Evacuation \sep Traffic Prediction \sep Graph Learning \sep Graph Transformer
\end{keyword}
\end{frontmatter}

\section{Introduction}\label{sec:intro}
Wildfires have become a persistent threat to the urban regions of the United States \citep{McConnell2026}. The continued development in the wildland--urban interface has increased the number of residents, homes, and infrastructure facilities exposed to wildfire hazards \citep{RadeloffEtAl2018}. Evacuation is one of the critical aspects of wildfire risk management. However, wildfire evacuation is difficult to manage because fire spread, warning information, household response, roadway availability, traffic conditions, etc. can change within a very short period. Unlike planned evacuations, wildfire evacuations often involve substantial uncertainty regarding the timing, location, and severity of the hazard. Rapid changes in fire behavior and weather conditions may require emergency management authorities to issue evacuation orders with limited advance notice or modify existing evacuation zones as the fire progresses. Consequently, a large number of residents may attempt to evacuate simultaneously, generating sudden increases in traffic demand on the surrounding transportation network.

The sudden increases in evacuation demand can significantly affect the overall traffic operations. Traffic conditions directly affect whether evacuees can leave threatened areas before evacuation routes become congested or unsafe. The resulting evacuation traffic interacts with existing background traffic, potentially leading to severe congestion on major evacuation corridors and surrounding roadways. Emergency-response vehicles, road closures, wildfire smoke, and uncertain route choices can further complicate traffic operations during this critical period \citep{IntiniEtAl2019,Kuligowski2021}. In particular, road closures and restrictions caused by advancing fire conditions may reduce the number of available evacuation routes, forcing additional traffic onto the remaining roadway network. Moreover, evacuees from different communities may attempt to use the same major corridors, creating localized bottlenecks and congestion that can propagate to neighboring roadway segments. Such conditions can delay evacuees attempting to leave threatened areas and restrict access for firefighters and other emergency responders traveling toward affected communities. Understanding how traffic conditions evolve across the transportation network is essential for supporting timely evacuation management decisions and maintaining safe and efficient evacuation operations. 

The Palisades and Eaton Fires demonstrated the scale and urgency of this problem. Both fires began in Los Angeles County on January 7, 2025. The Palisades Fire started near Pacific Palisades at approximately 10:30 a.m., and the Eaton Fire started near Altadena and Pasadena at approximately 6:18 p.m. Official records indicate that the Palisades Fire burned 23,448 acres, destroyed 6,845 structures, and caused 12 civilian fatalities. The Eaton Fire burned 14,021 acres, destroyed 9,419 structures, and caused 19 civilian fatalities \citep{CALFIREPalisades2025,CALFIREEaton2025}. Mobility analysis of the two fires identified substantial spatial variation in population displacement. Approximately 40\% of the observed evacuations occurred during the first day of fire \citep{JiangEtAl2026}. The simultaneous occurrence of the two fires produced a rapidly changing traffic environment across a large and interconnected metropolitan network. To better manage this evacuation traffic, emergency managers therefore need more than an estimate of the population that may evacuate. They also need location-specific forecasts of traffic flow across the roadway network. Such forecasts can support staged evacuations, traffic-signal control, ramp management, route guidance, traveler information, and the allocation of law-enforcement personnel.

However, predicting wildfire evacuation traffic is a challenging task because evacuation traffic characteristics can differ substantially from regular traffic. During the 2019 Kincade Fire, average vehicle speeds were approximately 3.5 km/h lower under evacuation conditions, and estimated roadway capacity decreased by about 5\% \citep{RohaertEtAl2023Kincade}. Analysis of the 2020 Glass Fire also identified reductions in traffic speed and roadway capacity during evacuation \citep{RohaertEtAl2023Glass}. For the 2020 Silverado Fire, vehicle speeds decreased by approximately 7.5 km/h, and estimated capacity on Interstate 5 decreased by 14.2\% compared with regular conditions \citep{JanfeshanaraghiEtAl2025}. The magnitude of these changes differed across the three events. This variation indicates that traffic conditions during a wildfire is considerably different than regular period traffic.

Another challenge is that detailed traffic observations during a wildfire event are rarely available before an event occurs. A model may therefore need to predict traffic during an evacuation even when it has been trained only on regular, pre-event traffic. Similar to the real-time traffic prediction models for hurricane evacuations \citep{RahmanHasan2023,RashidEtAl2025}, providing information about wildfire evacuation and fire progression rate may be beneficial for wildfire traffic prediction models. However, in practice, fire perimeters, road-closure information, evacuation orders, incident reports, population movement data, and other event-specific inputs may be incomplete, delayed, or unavailable in a form that can be directly incorporated into a network-wide real-time prediction system. It is therefore important to determine whether a model trained using routinely available traffic data can generalize to the abrupt and spatially uneven traffic conditions observed during an actual wildfire evacuation.

Although there are many wildfire evacuation studies focusing on household decisions, evacuation demand, route choice, clearance time, simulation-based planning or emerging mobility data \citep{CovaJohnson2002,WahlqvistEtAl2021,GrajduraEtAl2022,MaLee2025,ZhaoEtAl2022,JiangEtAl2026}, the literature on wildfire traffic prediction models remain very limited. \citet{HouEtAl2023} developed a support vector regression model for daily traffic speed and flow using historical fire, weather, and traffic data. \citet{ZhangEtAl2024} proposed a situational-aware multi-graph recurrent model to forecast census-tract-level travel demand during the Kincade Fire. These studies established the feasibility of data-driven model for wildfire evacuation traffic prediction. However, daily traffic prediction is not useful for real-time traffic operations. Additionally, travel demand at census-tract level may not provide accurate information about real-time traffic in a roadway location. Hence, these models are not suitable for network-wide detector-level hourly traffic prediction in real time.

Graph learning may be a suitable framework to solve this problem. As traffic observed at one detector may be influenced by traffic conditions at upstream and downstream detectors through directed roadway connections, the transportation network can be reliably represented as a graph \citep{RahmanHasan2023}. Graph-based traffic models have successfully represented spatial and temporal dependence under regular traffic conditions \citep{LiEtAl2018DCRNN,YuEtAl2018STGCN}. Graph learning has also been applied to network-wide hurricane evacuation traffic prediction, where a dynamic graph convolutional long short-term memory model produced forecasts up to six hours ahead \citep{RahmanHasan2023}. However, traffic prediction in wildfire evacuations presents a different generalization problem. Wildfires can begin with limited warning and produce localized, rapidly changing demand across an urban network. A model trained on regular traffic must recognize these changes from recent traffic observations and determine how relevant information should be transferred between connected detectors.

To address these problems, in this study, we develop a Graph Transformer--Long Short-Term Memory (GT-LSTM) model for network-wide real-time traffic prediction and evaluate it for the evacuation period of the Palisades and Eaton Fires. A roadway graph is constructed using detector-level information from 1,820 mainline traffic detector stations in California Department of Transportation District 7. At each input hour, a multi-head Graph Transformer (GT) processes the detector features and assigns attention to the neighboring detectors. The Gaussian distance weight is supplied to the transformer as an edge attribute so that the spatial representation reflects both current node conditions and roadway proximity. A spatial fusion gate then controls the amount of graph-transformer information transferred to each detector at each hour. The fusion gate limits the influence of uninformative neighboring conditions while retaining spatial information only when it improves prediction. The gated graph-transformer representation is combined with the original local detector features and a learnable station embedding. A Long Short-Term Memory (LSTM) module then learns the temporal progression of the combined representation. The previous six hours of data are used to predict traffic flow for the next six hours at 1-hour intervals.

The model is trained using regular period hourly traffic observations from October 1 through December 31, 2024 and is evaluated during the peak wildfire evacuation period from 11:00 a.m. on January 7 to 11:00 p.m. on January 8, 2025 without any event-specific retraining or fine-tuning. The test period covers the onset and peak evacuation activity of both the Palisades and Eaton Fires. The model does not depend on fire perimeters, fire distance, evacuation zones, evacuation orders, road closures, incident indicators, smoke conditions, population displacement, or other wildfire-specific variables. Instead, the model uses routinely available detector-level traffic, and roadway features and evaluates whether spatial and temporal patterns learned from regular traffic can support forecasting when the same network experiences an unseen wildfire evacuation conditions. 

The main contributions of this study are as follows:

\begin{itemize}
\item It synthesizes data from traffic detectors and represents the roadway network as a graph. No previous study considered a graph-based network-wide approach to predict hourly traffic flow during wildfire evacuation. 

\item It develops a Graph Transformer-Long Short-term Memory (GT-LSTM) framework that generates network-wide real-time traffic-flow forecasts up to 6-hours in advance during the Palisades and Eaton Fire period without relying on wildfire-specific input variables or model retraining.

\item It evaluates the generalizability of the proposed GT-LSTM model across different evacuation corridors and indicates potential congestion hotspots across the network.
\end{itemize}

\section{Study Area and Data Preparation}
\label{sec:study_area_data}

The Palisades and Eaton fires occurred in Los Angeles County, California, which is located within District 7 of the California Department of Transportation (Caltrans). Therefore, the Caltrans District 7 transportation network was selected as the study area. Detector-level hourly traffic data were collected from the Caltrans Performance Measurement System (PeMS) \citep{caltrans_pems}. The dataset includes hourly traffic volume, speed, and occupancy measurements.

\begin{figure}[htbp]
    \centering
    \includegraphics[width=0.70\textwidth]{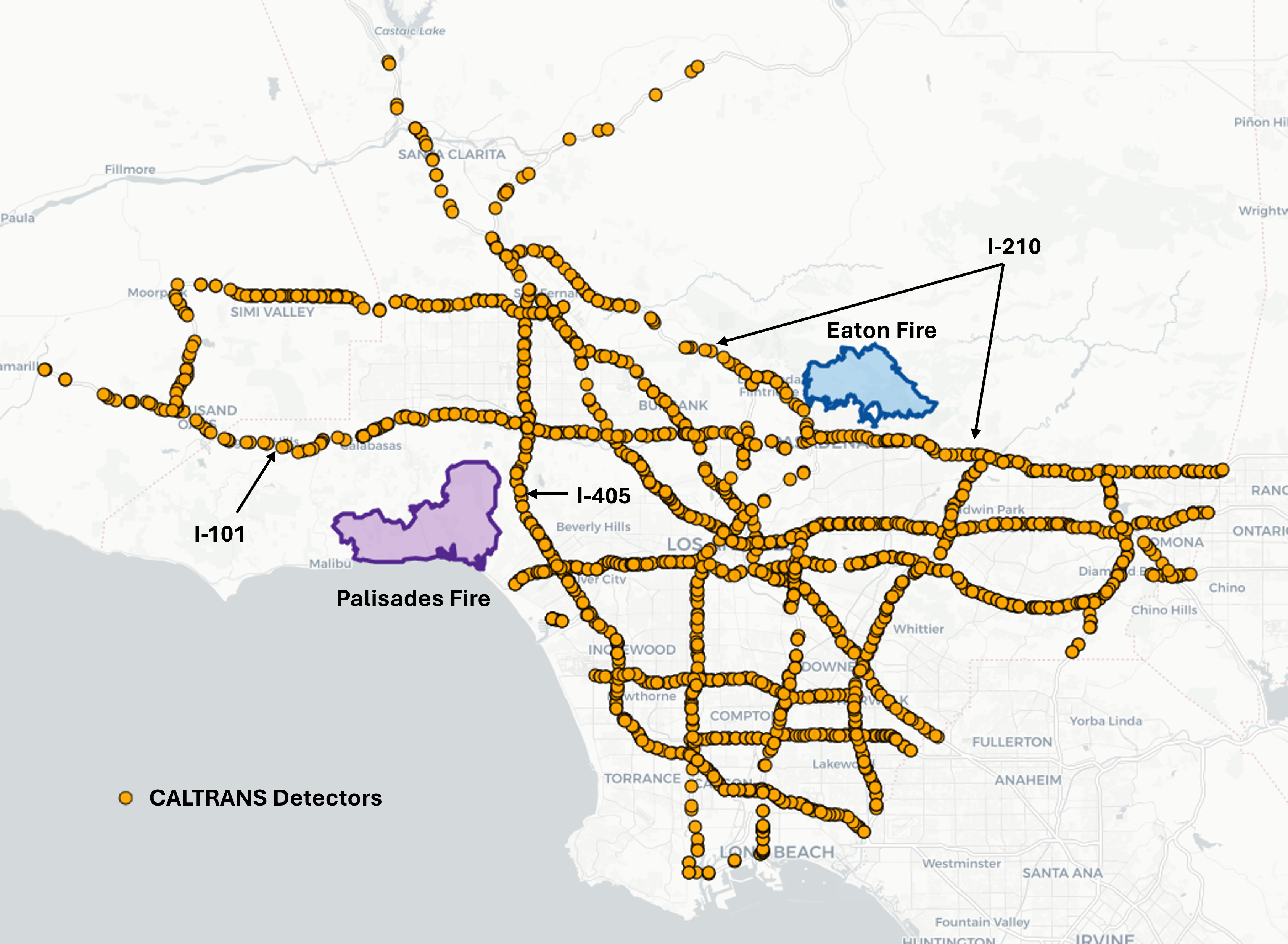}
    \caption{Traffic detector distribution in Caltrans District 7 and the Palisades and Eaton wildfire areas.}
    \label{fig:cal_d7}
\end{figure}

Figure~\ref{fig:cal_d7} presents the spatial distribution of traffic detectors across the District 7 network, together with the final fire perimeters of the Palisades and Eaton fires. Evacuation traffic originated near the fire-affected areas and propagated through multiple roadways across the regional transportation network. US-101 and I-405 are the two major highways closest to the Palisades fire, while I-210 is the major highway closest to the Eaton fire. Therefore, evacuees from the affected areas were likely to use these highways when traveling away from the fire zones.

To train the model under regular traffic conditions, hourly traffic data were collected for the District 7 network from October 1 to December 31, 2024. The Palisades fire began on January 7, 2025, at approximately 10:30 a.m., while the Eaton fire began later that day at approximately 6:18 p.m. To capture the traffic impacts associated with both wildfire evacuations, the period from 11:00 a.m. on January 7 to 11:00 p.m. on January 8, 2025, was selected as the model testing period. Although evacuation activity continued beyond this period, the most substantial network-wide traffic impacts were expected during the first one to two days after the fires began. Therefore, this 37-hour period was used to evaluate the model under wildfire evacuation conditions.

\begin{figure}[htbp]
    \centering
    \includegraphics[width=0.80\textwidth]{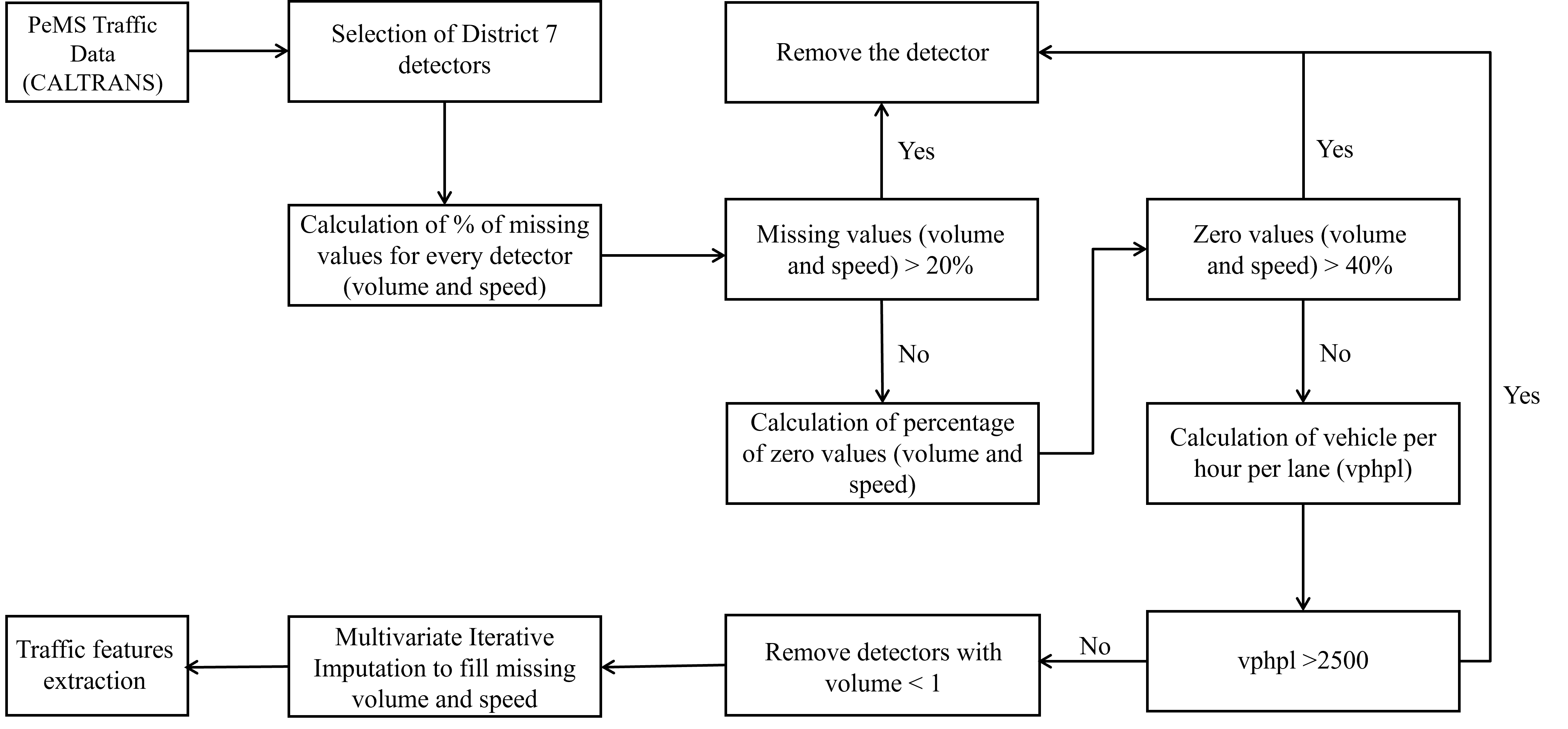}
    \caption{Data processing steps.}
    \label{fig:data_prep}
\end{figure}

Raw detector data may contain errors caused by detector malfunctions, adverse weather conditions, missing observations, duplicate records, and communication failures. In addition, congested stop-and-go traffic during evacuation may reduce detector accuracy and cause some vehicles to remain undetected \citep{RahmanHasan2023,rashid2024evacuation}. Directly using these raw observations could therefore introduce substantial errors during model training and evaluation. Similar to previous evacuation traffic prediction studies, an extensive data-cleaning procedure was applied to construct the final model input dataset, as summarized in Figure~\ref{fig:data_prep}.

First, the mainline detectors within the District 7 network were identified. The percentage of missing observations was calculated separately for traffic volume and speed at each detector. A detector was retained only when both variables contained no more than 20\% missing observations. The percentage of zero-valued observations was then calculated separately for volume and speed. Detectors were retained only when the proportion of zero values did not exceed 40\% for either variable.

Following previous hurricane evacuation studies \citep{RahmanHasan2023,rashid2024evacuation}, a maximum plausible traffic flow of 2,500 vehicles per hour per lane (vphpl) was adopted. Observations exceeding this threshold were considered potentially erroneous. Detectors containing such observations were removed from the dataset. Detectors with traffic flow consistently below one vehicle per hour were also removed because these observations indicated extremely sparse traffic and were unlikely to represent actual roadway conditions during the study period. After applying these criteria, the remaining missing traffic volume and speed observations were estimated using a multivariate iterative imputation method \citep{pedregosa2011scikit,RahmanHasan2023}.

To represent the periodic variation in daily traffic conditions, each 24-hour day was divided into six time periods: late night (12:00 a.m.--3:59 a.m.), early morning (4:00 a.m.--7:59 a.m.), morning (8:00 a.m.--11:59 a.m.), noon (12:00 p.m.--3:59 p.m.), evening (4:00 p.m.--7:59 p.m.), and night (8:00 p.m.--11:59 p.m.). These periods were represented using one-hot-encoded binary variables.

Additional historical traffic-profile features were included to capture recurring temporal patterns and short-term traffic propagation. These features consisted of the mean and standard deviation of traffic flow during the previous day and the mean and standard deviation of traffic flow during the corresponding preceding time period. The complete set of traffic-related input features is presented in Table~\ref{tab:features}.

\begin{table}[htbp]
\centering
\caption{Model input features}
\label{tab:features}
\renewcommand{\arraystretch}{1.0}
\begin{tabularx}{\linewidth}{X X}
\toprule
\textbf{Traffic Features} & \textbf{Temporal Indicators} \\
\midrule
Traffic volume (1-hour interval) & Hour-of-day (24 one-hot encoded) \\
Traffic speed (1-hour interval) & Weekday flag (binary) \\
Previous-day mean of volume &  Time-of-day period (6 one-hot encoded)\\
Previous-day std. of volume &  ~ \\
Previous-period mean of volume & ~ \\
Previous-period std. of volume & ~ \\
Number of lanes & ~ \\
\bottomrule
\end{tabularx}
\end{table}
\section{Methodology}
\label{sec:methodology}

\subsection{Problem Formulation}
\label{subsec:problem_formulation}

We represent the Caltrans District 7 roadway network as a directed graph $\mathcal{G}=(\mathcal{V},\mathcal{E},\mathbf{A})$, where $\mathcal{V}$ is the set of traffic detector stations, $\mathcal{E}$ is the set of directed roadway connections, and $\mathbf{A}$ is the distance-weighted adjacency matrix. The network contains $N=1820$ mainline detector stations and $1{,}778$ directed roadway connections. Each directed edge $(j,i)\in\mathcal{E}$ represents traffic movement from source detector station $j$ to downstream target detector station $i$.

At hour $t$, each station $i$ is represented by a feature vector $\mathbf{x}_{i,t}\in\mathbb{R}^{F}$, where $F$ is the number of features. The network-wide feature matrix is $\mathbf{X}_{t}
=
[
\mathbf{x}_{1,t},
\mathbf{x}_{2,t},
\ldots,
\mathbf{x}_{N,t}
]^{\top}
\in\mathbb{R}^{N\times F}$. The proposed Graph Transformer--Long Short-Term Memory (GT--LSTM) model uses the previous six hours of traffic features to predict traffic volume for the next six hours. For a prediction origin $t$, the model input can be defined as, $\mathcal{X}_{t}
=
[
\mathbf{X}_{t-l+1},
\mathbf{X}_{t-l+2},
\ldots,
\mathbf{X}_{t}
]
\in\mathbb{R}^{l\times N\times F}$. Here, $l$ is the dimension of the input horizon. The model target are defined as, $\mathbf{Y}_{t}
=
[
\mathbf{y}_{t+1},
\mathbf{y}_{t+2},
\ldots,
\mathbf{y}_{t+p}
]
\in\mathbb{R}^{N\times p}$. Here, $\mathbf{y}_{t+p}\in\mathbb{R}^{N}$ contains the observed traffic volumes at prediction horizon $p$. The traffic prediction task can be is expressed as
\begin{equation}
\hat{\mathbf{Y}}_{t}
=
f_{\boldsymbol{\Theta}}
(
\mathcal{X}_{t},\mathcal{G}
),
\label{eq:forecasting_function}
\end{equation}
where, $\hat{\mathbf{Y}}_{t}$ is the predicted traffic flow for time $t$. $f_{\boldsymbol{\Theta}}(\cdot)$ denotes the GT--LSTM model with learnable parameters $\boldsymbol{\Theta}$.

\subsection{Input and Target Feature Processing}
Each input feature was standardized by global z-score standardization (Equation~\ref{eq:input_standardization}) using the mean and standard deviation estimated from the training data. 
\begin{equation}
\tilde{x}_{i,t,f}
=
\frac{x_{i,t,f}-\mu_f^{x}}
{\sigma_f^{x}},
\qquad f=1,\ldots,F,
\label{eq:input_standardization}
\end{equation}
where $\mu_f^{x}$ and $\sigma_f^{x}$ are the network-wide training mean and standard deviation of feature $f$.

As traffic-volume levels differ substantially among detectors, the target variable was standardized separately for each detector using detector-specific z-score standardization (Equation~\ref{eq:target_standardization}). 
\begin{equation}
\tilde{y}_{i,t}
=
\frac{y_{i,t}-\mu_i^{y}}
{\sigma_i^{y}},
\label{eq:target_standardization}
\end{equation}
where $\mu_i^{y}$ and $\sigma_i^{y}$ are the mean and standard deviation of the observed volumes at station $i$. The station-specific transformation prevents high-volume detectors from dominating the loss function.

\subsection{Gaussian-Weighted Roadway Graph}
\label{subsec:gaussian_graph}

The graph connects sequential detector stations according to roadway direction. Hence, spatial information follow the original source-to-target direction of traffic movement. The distance associated with edge $(j,i)$ was obtained from the   roadway distance or post-mile distance from raw data, whichever was available. When both attributes were unavailable, the Haversine distance between the two detector coordinates was used. Let $d_{ji}$ denote the distance in miles from source station $j$ to target station $i$. We perform a Gaussian transformation for each edge distance $d_{ji}$, so that short distance correspond to stronger weight during message passing. The weighted adjacency matrix can be defined as,
\begin{equation}
A_{ji}
=
\begin{cases}
\displaystyle
w_{ji}=\exp\left[-\left(\dfrac{d_{ji}}{\sigma}\right)^2\right],
&
(j,i)\in\mathcal{E}
\ \text{and}\
w_{ji}\geq\tau,\\[8pt]
0,
&
\text{otherwise}.
\end{cases}
\label{eq:thresholded_adjacency}
\end{equation}
where $\sigma$  is a a distance-decay bandwidth. It controls how rapidly the spatial weight decreases as the distance between two connected detectors increases. $\tau$ is the minimum Gaussian weight required to retain an edge. The threshold removes edges with very weak spatial relationships and prevents negligible long-distance connections from introducing noise into the message-passing process. This formulation is asymmetric because $A_{ji}$ is defined only when the directed edge $(j,i)$ exists in graph $\mathcal{G}$.

The initial value of $\sigma$ was set to the median positive edge distance, $\sigma_{\mathrm{base}}
=
\operatorname{median}
\{d_{ji}:d_{ji}>0,\,(j,i)\in\mathcal{E}\}.
\label{eq:base_sigma}$. The median was used because it provides a representative network distance while being less sensitive than the mean to unusually long roadway connections. The resulting base bandwidth was approximately $0.46$ miles. Candidate values of $\sigma$ were constructed using multipliers $\{1.0,1.5,2.0\}$, while the candidate thresholds $\tau$ were $\{0,0.05\}$. The final parameters were selected using validation RMSE. The selected configuration retained 1,727 out of the 1,778 directed roadway edges. The scalar Gaussian weight $\mathbf{e}_{ji}=[A_{ji}]\in\mathbb{R}^{1}$ was then supplied to the Graph Transformer as an edge attribute.

\subsection{Graph Transformer--LSTM Architecture}
\label{subsec:gt_lstm}

\begin{figure}[htbp]
    \centering
    \includegraphics[width=\textwidth]{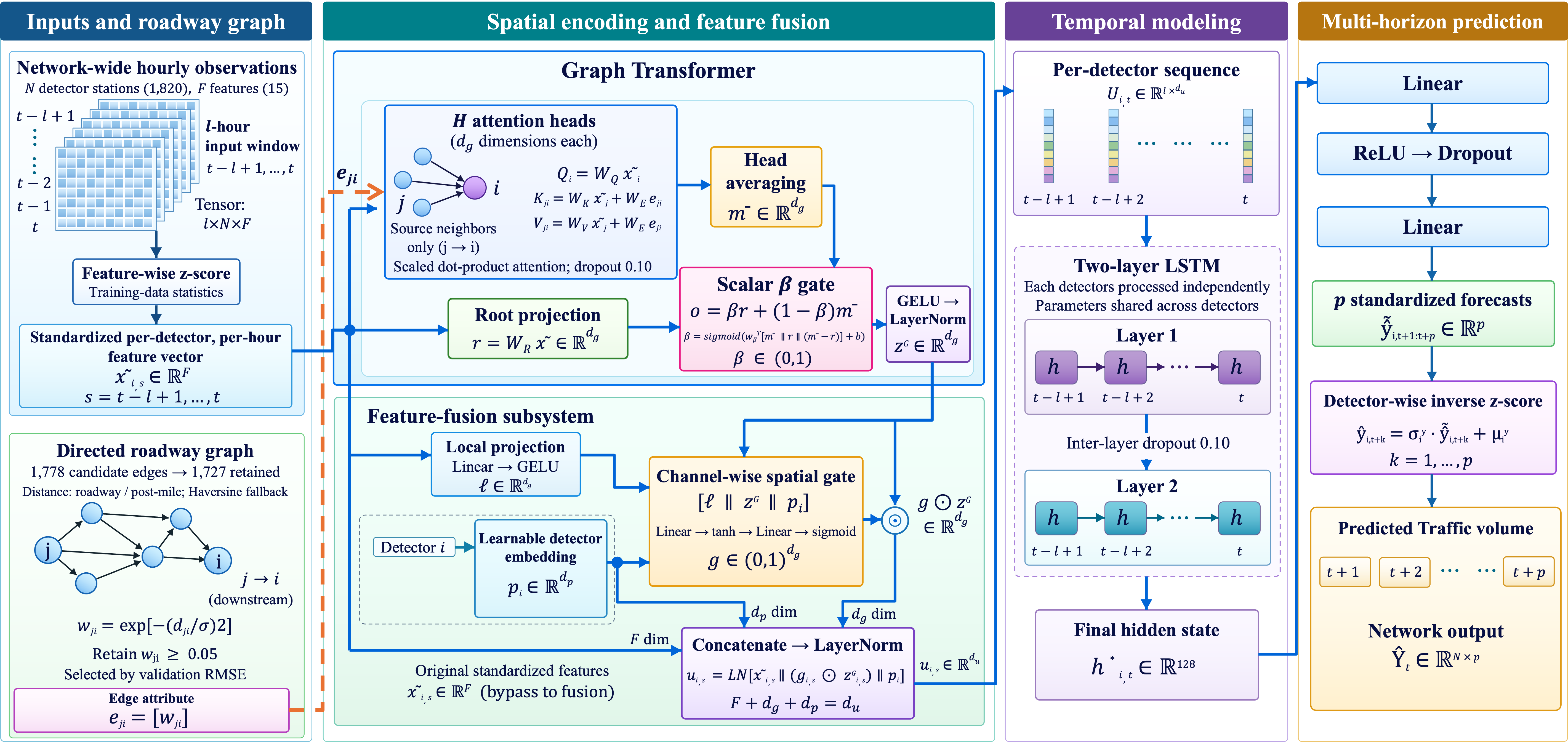}
    \caption{Overview of the proposed GT--LSTM framework.}
    \label{fig:model}
\end{figure}

The GT--LSTM architecture contains four main components: a multi-head Graph Transformer, a learnable station embedding, a spatial fusion gate, and a two-layer LSTM. Figure~\ref{fig:model} shows the overview of the proposed model. At every input hour, the Graph Transformer learns the interactions between directly connected detectors. The spatial representations are subsequently processed by the LSTM to learn temporal dependence.

\subsubsection{Multi-Head Graph Transformer}

A single Graph Transformer layer with $H$ attention heads was applied independently to each of the six input-hour graphs. Let $\mathcal{N}^{-}(i)
=
\{j:(j,i)\in\mathcal{E}\}$ denote the set of stations with directed edges entering station $i$. Thus, station $i$ receives messages only from its directed source neighbors.

For attention head $h$, the query $\mathbf{q}_{i,t}^{(h)}$, key $\mathbf{k}_{ji,t}^{(h)}$, and value $\mathbf{v}_{ji,t}^{(h)}$ representations are calculated as,
\begingroup
\setlength{\jot}{2pt}
\begin{align}
\mathbf{q}_{i,t}^{(h)}
&=
\mathbf{W}_{Q}^{(h)}
\tilde{\mathbf{x}}_{i,t}
\label{eq:query}
\\
\mathbf{k}_{ji,t}^{(h)}
&=
\mathbf{W}_{K}^{(h)}
\tilde{\mathbf{x}}_{j,t}
+
\mathbf{W}_{E}^{(h)}
\mathbf{e}_{ji}
\label{eq:key}
\\
\mathbf{v}_{ji,t}^{(h)}
&=
\mathbf{W}_{V}^{(h)}
\tilde{\mathbf{x}}_{j,t}
+
\mathbf{W}_{E}^{(h)}
\mathbf{e}_{ji}
\label{eq:value}
\end{align}

\endgroup
where $\tilde{\mathbf{x}}_{i,t}\in\mathbb{R}^{F}$ is the standardized input vector, $\mathbf{e}_{ji}$ is the Gaussian edge attribute, and each attention head has hidden dimension $d_g$. The same learned edge transformation is added to both the key and value representations, similar to the Graph Transformer operator developed by \citep{ShiEtAl2021}. The attention coefficient from source station $j$ to target station $i$ is shown in Equation~\ref{eq:attention}.
\begin{equation}
\alpha_{ji,t}^{(h)}
=
\frac{
\exp[
(\mathbf{q}_{i,t}^{(h)})^{\top}
\mathbf{k}_{ji,t}^{(h)}
/\sqrt{d_g}
]
}{
\displaystyle
\sum_{r\in\mathcal{N}^{-}(i)}
\exp[
(\mathbf{q}_{i,t}^{(h)})^{\top}
\mathbf{k}_{ri,t}^{(h)}
/\sqrt{d_g}
]
}.
\label{eq:attention}
\end{equation}

Attention dropout with a rate of 0.10 was applied to the normalized coefficients during training. The message produced by each attention head $h$ and the resulting multi-head spatial message, obtained by averaging the outputs of all attention heads are expressed by Equation~\ref{eq:head_message} and \ref{eq:head_average} respectively:
\begin{equation}
\mathbf{m}_{i,t}^{(h)}
=
\sum_{j\in\mathcal{N}^{-}(i)}
\operatorname{Dropout}
(
\alpha_{ji,t}^{(h)}
)
\mathbf{v}_{ji,t}^{(h)}.
\label{eq:head_message}
\end{equation}

\begin{equation}
\overline{\mathbf{m}}_{i,t}
=
\frac{1}{H}
\sum_{h=1}^{H}
\mathbf{m}_{i,t}^{(h)}
\in\mathbb{R}^{d_g}.
\label{eq:head_average}
\end{equation}

Although spatially connected detectors often exhibit similar traffic patterns, their traffic conditions can sometime differ substantially. During wildfire evacuations, these differences may become more pronounced due to uneven evacuation demand, route diversions, incidents, and road closures. For example, a detector located immediately upstream of a congested roadway segment may experience substantially different traffic conditions from a neighboring detector located downstream of the bottleneck. Under such circumstances, the target detector's own traffic information may provide a more representative description of its current traffic condition than the aggregated information received from neighboring detectors. Previous research has demonstrated that aggregating information from neighboring nodes with dissimilar features can reduce the effectiveness of graph neural networks, highlighting the importance of preserving local node representations \citep{ZhuEtAl2020}. Hence, a transformed root-node representation is also retained to preserve the current local condition of the target detector (see Equation~\ref{eq:root_representation}).
\begin{equation}
\mathbf{r}_{i,t}
=
\mathbf{W}_{R}
\tilde{\mathbf{x}}_{i,t}.
\label{eq:root_representation}
\end{equation}
Here, $\mathbf{W}_{R}\in\mathbb{R}^{d_g\times F}$ is a learnable linear transformation matrix. The Graph Transformer learns a scalar beta gate that balances the root-node representation and the aggregated neighbor message:
\begin{equation}
\beta_{i,t}
=
\operatorname{sigmoid}
\{
\mathbf{w}_{\beta}^{\top}
[
\overline{\mathbf{m}}_{i,t}
\mathbin\Vert
\mathbf{r}_{i,t}
\mathbin\Vert
(\overline{\mathbf{m}}_{i,t}-\mathbf{r}_{i,t})
]
+
b_{\beta}
\}.
\label{eq:beta_gate}
\end{equation}
where $\Vert$ denotes vector concatenation. Equation~\ref{eq:beta_gate} contains the neighborhood representation, the local root-node representation, and their element-wise difference. The learnable vector $\mathbf{w}_{\beta}\in\mathbb{R}^{d_g}$ assigns importance to these components. $b_{\beta}\in\mathbb{R}$ is a learnable bias term. The sigmoid function constrains the resulting gate value to the interval $(0,1)$. Including the difference term $\overline{\mathbf{m}}_{i,t}-\mathbf{r}_{i,t}$ allows the gate to measure the level of agreement or disagreement between the traffic condition at the target detector and the information received from its neighboring detectors. The gate allows the model to adaptively determine whether the local detector condition or the surrounding roadway condition should receive greater emphasis.

The Graph Transformer output before activation is as follows,
\begin{equation}
\mathbf{o}_{i,t}^{G}
=
\beta_{i,t}\mathbf{r}_{i,t}
+
(1-\beta_{i,t})
\overline{\mathbf{m}}_{i,t}.
\label{eq:transformer_residual}
\end{equation}
Here, $\mathbf{o}_{i,t}^{G}\in\mathbb{R}^{d_g}$ is the fused Graph Transformer output before nonlinear activation. When $\beta_{i,t}$ is close to one, the model places greater emphasis on the local root-node representation $\mathbf{r}_{i,t}$. When $\beta_{i,t}$ is close to zero, the model relies more heavily on the aggregated neighborhood message $\overline{\mathbf{m}}_{i,t}$. This adaptive fusion prevents the neighborhood aggregation from excessively smoothing the detector-specific traffic condition. At the same time, it allows the model to use upstream spatial information when neighboring traffic patterns are informative. 

The final spatial representation is obtained using the Gaussian Error Linear Unit (GELU) activation function and layer normalization:
\begin{equation}
\mathbf{z}_{i,t}^{G}
=
\operatorname{LayerNorm}
[
\operatorname{GELU}
(
\mathbf{o}_{i,t}^{G}
)
]
\in\mathbb{R}^{d_g}.
\label{eq:transformer_output}
\end{equation}

\subsubsection{Station Embedding and Spatial Fusion Gate}

A learnable station embedding was used to represent persistent location-specific characteristics that may not be fully described by the observed features. For station $i$, the embedding is
\begin{equation}
\mathbf{p}_{i}
=
\mathbf{E}_{s}[i]
\in\mathbb{R}^{d_p},
\label{eq:station_embedding}
\end{equation}
where $\mathbf{E}_{s}\in\mathbb{R}^{N\times d_p}$ is a trainable embedding matrix. $d_p$ denote the station-embedding dimension. The same embedding is used for all hours associated with a station.

The local node features are first projected into the same $d_g$-dimensional space as the Graph Transformer output:
\begin{equation}
\boldsymbol{\ell}_{i,t}
=
\operatorname{GELU}
(
\mathbf{W}_{L}
\tilde{\mathbf{x}}_{i,t}
+
\mathbf{b}_{L}
)
\in\mathbb{R}^{d_g}.
\label{eq:local_projection}
\end{equation}

A channel-wise spatial gate determines how much of the Graph Transformer representation is passed to the temporal model:
\begin{equation}
\mathbf{g}_{i,t}
=
\operatorname{sigmoid}
\{
\mathbf{W}_{g,2}
\tanh
[
\mathbf{W}_{g,1}
(
\boldsymbol{\ell}_{i,t}
\mathbin\Vert
\mathbf{z}_{i,t}^{G}
\mathbin\Vert
\mathbf{p}_{i}
)
+
\mathbf{b}_{g,1}
]
+
\mathbf{b}_{g,2}
\},
\label{eq:spatial_gate}
\end{equation}
where $\mathbf{g}_{i,t}\in(0,1)^{d_g}$. The second gate-layer weights were initialized to zero, and its bias was initialized to $-3$. Therefore, the initial gate value was $\operatorname{sigmoid}(-3)\approx0.047$, which allowed the model to begin training near the local-feature pathway and gradually increase the contribution of spatial information where supported by the data.

The gated spatial representation is
\begin{equation}
\tilde{\mathbf{z}}_{i,t}^{G}
=
\mathbf{g}_{i,t}
\odot
\mathbf{z}_{i,t}^{G},
\label{eq:gated_spatial_representation}
\end{equation}
where $\odot$ denotes element-wise multiplication. The input passed to the temporal module is
\begin{equation}
\mathbf{u}_{i,t}
=
\operatorname{LayerNorm}
[
\tilde{\mathbf{x}}_{i,t}
\mathbin\Vert
\tilde{\mathbf{z}}_{i,t}^{G}
\mathbin\Vert
\mathbf{p}_{i}
]
\in\mathbb{R}^{d_u}.
\label{eq:fused_representation}
\end{equation}
where \(d_u\) denotes the dimension of the fused temporal representation and is defined as, $d_u
=F+d_g+d_p$. The standardized input features are directly included in the fusion process so that the subsequent temporal module can learn from the original station-level observations alongside the spatially aggregated representation generated by the graph transformer.

\subsubsection{Temporal Modeling with LSTM}

For each station, the six fused representations are arranged as
\begin{equation}
\mathbf{U}_{i,t}
=
[
\mathbf{u}_{i,t-5},
\mathbf{u}_{i,t-4},
\ldots,
\mathbf{u}_{i,t}
]
\in\mathbb{R}^{6\times d_u}.
\end{equation}
Each station sequence is processed independently by a two-layer LSTM with $d_h$ hidden units per layer \citep{HochreiterSchmidhuber1997}. The LSTM parameters are shared among all detector stations. For time step $s$ and LSTM layer $\ell$, the input, forget, output, and candidate-cell equations are follows,
\begingroup
\setlength{\jot}{2pt}
\begin{align}
\boldsymbol{\iota}_{s}^{(\ell)}
&=
\operatorname{sigmoid}
\left(
\mathbf{W}_{\iota}^{(\ell)}
\mathbf{q}_{s}^{(\ell)}
+
\mathbf{U}_{\iota}^{(\ell)}
\mathbf{h}_{s-1}^{(\ell)}
+
\mathbf{b}_{\iota}^{(\ell)}
\right)
\label{eq:lstm_input_gate}
\\
\mathbf{f}_{s}^{(\ell)}
&=
\operatorname{sigmoid}
\left(
\mathbf{W}_{f}^{(\ell)}
\mathbf{q}_{s}^{(\ell)}
+
\mathbf{U}_{f}^{(\ell)}
\mathbf{h}_{s-1}^{(\ell)}
+
\mathbf{b}_{f}^{(\ell)}
\right)
\label{eq:lstm_forget_gate}
\\
\mathbf{o}_{s}^{(\ell)}
&=
\operatorname{sigmoid}
\left(
\mathbf{W}_{o}^{(\ell)}
\mathbf{q}_{s}^{(\ell)}
+
\mathbf{U}_{o}^{(\ell)}
\mathbf{h}_{s-1}^{(\ell)}
+
\mathbf{b}_{o}^{(\ell)}
\right)
\label{eq:lstm_output_gate}
\\
\tilde{\mathbf{c}}_{s}^{(\ell)}
&=
\tanh
\left(
\mathbf{W}_{c}^{(\ell)}
\mathbf{q}_{s}^{(\ell)}
+
\mathbf{U}_{c}^{(\ell)}
\mathbf{h}_{s-1}^{(\ell)}
+
\mathbf{b}_{c}^{(\ell)}
\right)
\label{eq:lstm_candidate}
\end{align}

\endgroup

The cell and hidden states are updated using Equation~\ref{eq:lstm_cell} and \ref{eq:lstm_hidden}.
\begingroup
\setlength{\jot}{2pt}
\begin{align}
\mathbf{c}_{s}^{(\ell)}
&=
\mathbf{f}_{s}^{(\ell)}
\odot
\mathbf{c}_{s-1}^{(\ell)}
+
\boldsymbol{\iota}_{s}^{(\ell)}
\odot
\tilde{\mathbf{c}}_{s}^{(\ell)}
\label{eq:lstm_cell}
\\
\mathbf{h}_{s}^{(\ell)}
&=
\mathbf{o}_{s}^{(\ell)}
\odot
\tanh
\left(
\mathbf{c}_{s}^{(\ell)}
\right)
\label{eq:lstm_hidden}
\end{align}

\endgroup
For the first LSTM layer, $\mathbf{q}_{s}^{(1)}=\mathbf{u}_{i,s}$. For the second layer,
\begin{equation}
\mathbf{q}_{s}^{(2)}
=
\operatorname{Dropout}
(
\mathbf{h}_{s}^{(1)}
)
\end{equation}
where the inter-layer dropout rate is considered 0.10. The final hidden state of the second layer after the sixth input hour, denoted by $\mathbf{h}_{i,t}^{*}\in\mathbb{R}^{d_h}$, summarizes the spatial and temporal information for station $i$.

\subsubsection{Multi-Horizon Output Layer}

The final LSTM state is passed through a two-layer feedforward prediction head:
\begin{equation}
\mathbf{a}_{i,t}
=
\operatorname{ReLU}
(
\mathbf{W}_{1}
\mathbf{h}_{i,t}^{*}
+
\mathbf{b}_{1}
),
\qquad
\mathbf{a}_{i,t}\in\mathbb{R}^{d_o}
\label{eq:output_hidden}
\end{equation}
where, \(\mathbf{W}_{1}\in\mathbb{R}^{d_o\times d_h}\) and \(\mathbf{b}_{1}\in\mathbb{R}^{d_o}\) are the weight matrix and bias vector of the first feedforward layer, respectively, and \(\mathbf{a}_{i,t}\) is the intermediate hidden representation. \(d_o\) is the hidden dimension of the prediction head.

The standardized multi-horizon prediction vector is then obtained as,
\begin{equation}
\hat{\tilde{\mathbf{y}}}_{i,t}
=
\mathbf{W}_{2}
\operatorname{Dropout}
(
\mathbf{a}_{i,t}
)
+
\mathbf{b}_{2},
\qquad
\hat{\tilde{\mathbf{y}}}_{i,t}
\in\mathbb{R}^{p}.
\label{eq:multi_horizon_output}
\end{equation}
where \(\mathbf{W}_{2}\in\mathbb{R}^{p\times d_o}\) and \(\mathbf{b}_{2}\in\mathbb{R}^{p}\) are the weight matrix and bias vector of the output layer, respectively. The output vector contains simultaneous predictions for horizons $t+1$ through $t+p$. Predictions are converted to their original traffic-volume units using the station-specific target statistics:
\begin{equation}
\hat{y}_{i,t+p}
=
\hat{\tilde{y}}_{i,t+p}
\sigma_i^{y}
+
\mu_i^{y},
\qquad p=1,\ldots,6.
\label{eq:inverse_target_scaling}
\end{equation}

\subsection{Model Training and Hyperparameter Selection}
\label{subsec:model_training}
All input--output sequences between October 1 and December 31, 2024 were ordered chronologically. The first 90\% of the rolling sequences were used for training, and the remaining 10\% were used for validation. This produced 1,977 training sequences and 220 validation sequences. The test targets extended from 11:00 a.m. on January 7 to 11:00 p.m. on January 8, 2025.

The model used an input feature dimension of $F=15$, a graph representation dimension of $d_g=32$, and a detector embedding dimension of $d_p=16$. The fused input dimension supplied to the LSTM was $d_u=F+d_g+d_p=63$. Let $M_{b,i,h}$ indicate whether the target for sample $b$, station $i$, and horizon $h$ is observed. The model was trained by minimizing the mean squared error in standardized target units (see Equation~\ref{eq:masked_mse}).
\begin{equation}
\mathcal{L}
=
\frac{
\displaystyle
\sum_{b=1}^{B}
\sum_{i=1}^{N}
\sum_{p=1}^{6}
M_{b,i,p}
(
\hat{\tilde{y}}_{b,i,p}
-
\tilde{y}_{b,i,p}
)^2
}{
\displaystyle
\sum_{b=1}^{B}
\sum_{i=1}^{N}
\sum_{p=1}^{6}
M_{b,i,p}
}.
\label{eq:masked_mse}
\end{equation}

The model was trained using the Adam optimizer with an initial learning rate of $5\times10^{-4}$, weight decay of $10^{-5}$, and batch size of 8. Gradient norms were clipped at 5.0. A reduce-on-plateau scheduler decreased the learning rate by a factor of 0.5 when validation performance stopped improving, with a minimum learning rate of $10^{-6}$. Training was conducted for a maximum of 150 epochs. Early stopping was applied when the validation RMSE did not improve for 25 consecutive epochs. The model configuration contained 279,908 trainable parameters.

\subsection{Performance Evaluation}
\label{subsec:performance_evaluation}

Model performance was evaluated using root mean square error (RMSE), mean absolute error (MAE), symmetric mean absolute percentage error (SMAPE), and the coefficient of determination ($R^2$). For $K$ valid observations, these measures are defined as follows,
\begingroup
\setlength{\jot}{2pt}

\begin{align}
\operatorname{RMSE}
&=
\sqrt{
\frac{1}{K}
\sum_{k=1}^{K}
\left(
y_k-\hat{y}_k
\right)^2
}
\label{eq:rmse}
\\
\operatorname{MAE}
&=
\frac{1}{K}
\sum_{k=1}^{K}
\left|
y_k-\hat{y}_k
\right|
\label{eq:mae}
\\
\operatorname{SMAPE}
&=
\frac{100}{K}
\sum_{k=1}^{K}
\frac{
\left|
y_k-\hat{y}_k
\right|
}{
\left(
\left|y_k\right|
+
\left|\hat{y}_k\right|
\right)/2
}
\label{eq:smape}
\\
R^2
&=
1-
\frac{
\displaystyle
\sum_{k=1}^{K}
\left(
y_k-\hat{y}_k
\right)^2
}{
\displaystyle
\sum_{k=1}^{K}
\left(
y_k-\overline{y}
\right)^2
}
\label{eq:r_squared}
\end{align}

\endgroup
where,
 \( K \) is the total number of samples;
 \( y_{\text{k}} \) represents actual flow for \( k \)-th sample;
 \( \hat{y}_{\text{k}} \) represents predicted flow for \( k \)-th sample; and
 \( \overline{y} \) is the mean of the actual flows. Horizon-specific metrics were calculated separately for each one- through six-hour forecast horizon.

\section{Results}
We predicted the network-wide traffic flow during the evacuation period of Palisades and Eaton Fire. Table~\ref{tab:network_wide_wildfire_metrics} presents the network-wide prediction performance of the GT--LSTM model during test evacuation period. The model achieved its highest accuracy at the one-hour prediction horizon, with an RMSE of 291.7, an MAE of 154.7, a SMAPE of 6.41\%, and an $R^2$ of 0.98. It indicates that the model accurately reproduced short-term network-wide traffic conditions and explained approximately 98\% of the observed variation in traffic flow. Prediction errors increased gradually as the forecast horizon became longer. At the six-hour horizon, the RMSE increased to 594.6, the MAE increased to 292.6, and the SMAPE increased to 12.37\%. Nevertheless, across all prediction horizons, the model achieved an overall RMSE of 435.1, an MAE of 221.3, a SMAPE of 8.90\%, and an $R^2$ of 0.96. The overall MAE was approximately 6.1\% of the mean observed traffic flow, which indicates a relatively small average deviation when considered against the magnitude of network-wide traffic volumes. The increase in error with prediction horizon was expected because longer-term forecasts are affected by increasing uncertainty in evacuation demand, route choice, background traffic, and the propagation of traffic disturbances. However, the consistently high $R^2$ values demonstrate that the GT--LSTM model preserved strong predictive capability across the complete six-hour forecasting period.

\begin{table}[htbp]
\centering

\caption{Network-wide wildfire traffic prediction performance during the wildfire evacuation period (January 7, 11:00 a.m. to January 8, 11:00 p.m.)}
\label{tab:network_wide_wildfire_metrics}

\begingroup
\small
\setlength{\tabcolsep}{6pt}
\renewcommand{\arraystretch}{1.08}

\begin{tabular}{lcccc}
\toprule
\textbf{Prediction horizon}
& \textbf{RMSE}
& \textbf{MAE}
& \textbf{SMAPE (\%)}
& \textbf{$R^2$} \\
\midrule
1-hour & 291.7 & 154.7 & 6.41 & 0.98 \\
2-hour & 394.6 & 201.9 & 8.00 & 0.97 \\
3-hour & 473.4 & 234.0 & 9.28 & 0.96 \\
4-hour & 531.4 & 258.0 & 10.25 & 0.94 \\
5-hour & 566.0 & 276.0 & 11.24 & 0.93 \\
6-hour & 594.6 & 292.6 & 12.37 & 0.93 \\
\midrule
\textbf{Overall}
& \textbf{435.1}
& \textbf{221.3}
& \textbf{8.90}
& \textbf{0.96} \\
\bottomrule
\end{tabular}

\endgroup

\vspace{4pt}

\begin{minipage}{0.95\textwidth}
\footnotesize
\textit{Note.} Minimum flow = 39; maximum flow = 15,873;
mean flow = 3,649.40; median flow = 3,757.
\end{minipage}

\end{table}

\begin{figure}[htbp]
    \centering
    \includegraphics[
        width=0.65\textwidth
    ]{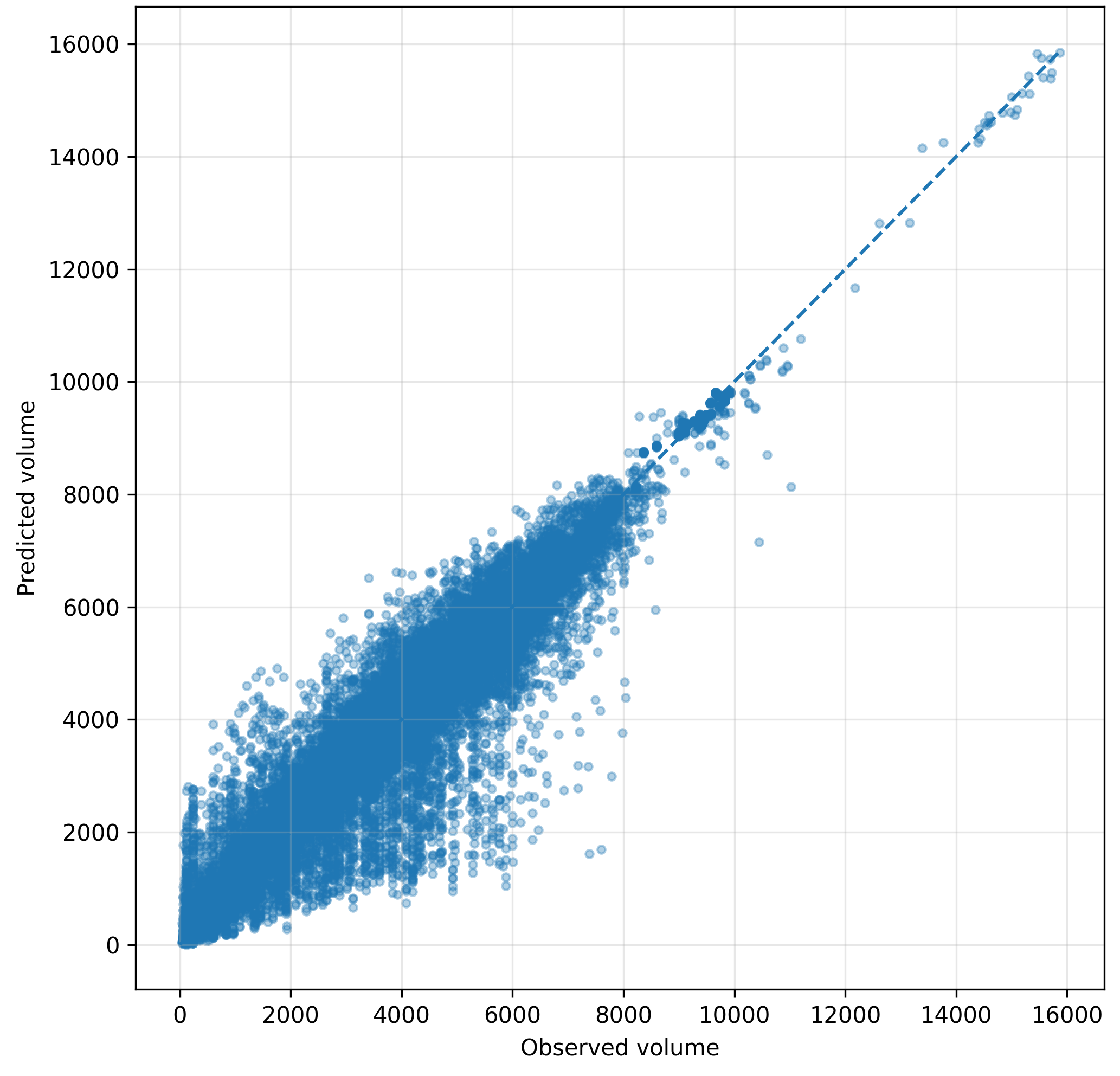}
    \caption{Comparison of observed and predicted traffic flow during the wildfire evacuation test period.}
    \label{fig:scatter}
\end{figure}

Figure~\ref{fig:scatter} further compares the observed and predicted traffic volumes for the complete test period. This result shows that the model did not simply reproduce average traffic conditions but was also able to represent substantial differences in flow among detector stations.

Table~\ref{tab:wildfire_baseline_model_comparison} compares the GT--LSTM model with state-of-the-art LSTM, convolutional neural network (CNN), CNN--LSTM, graph convolutional network (GCN), support vector regression (SVR), and GCN--LSTM models. The proposed GT--LSTM model achieved the lowest RMSE, MAE, and SMAPE at every prediction horizon. Among the non-graph temporal models, the LSTM achieved the best overall RMSE of 616.55, while the CNN achieved the lowest overall MAE and SMAPE of 354.72 and 14.67\%, respectively. The GCN--LSTM was the strongest baseline model. It achieved an overall RMSE of 447.01, and an $R^2$ of 0.96. The substantial improvement of GCN--LSTM over the individual GCN and LSTM models demonstrates the importance of jointly representing spatial and temporal traffic dependencies. Nevertheless, the GT--LSTM consistently outperformed the GCN--LSTM. Relative to GCN--LSTM, the proposed model reduced the overall RMSE by 2.66\%, the MAE by 3.94\%, and the SMAPE by 4.71\%.

\begin{table}[htbp]
\centering

\caption{Comparison of baseline and proposed models for wildfire traffic-flow prediction}
\label{tab:wildfire_baseline_model_comparison}

\begingroup
\scriptsize
\renewcommand{\arraystretch}{0.95}

\resizebox{\textwidth}{!}{%
\begin{tabular}{llcccc}
\toprule
\textbf{Model}
& \textbf{Horizon}
& \textbf{RMSE}
& \textbf{MAE}
& \textbf{SMAPE (\%)}
& \textbf{$R^2$} \\
\midrule

LSTM & 1-hour & 527.37 & 303.24 & 12.68 & 0.95 \\
 & 2-hour & 595.48 & 349.64 & 14.47 & 0.93 \\
 & 3-hour & 651.80 & 380.94 & 15.61 & 0.92 \\
 & 4-hour & 697.51 & 400.91 & 16.30 & 0.90 \\
 & 5-hour & 728.13 & 415.65 & 16.91 & 0.89 \\
 & 6-hour & 750.59 & 428.46 & 17.69 & 0.88 \\
 & Overall & 616.55 & 358.16 & 14.73 & 0.92 \\
\midrule

CNN & 1-hour & 530.03 & 295.15 & 12.32 & 0.94 \\
 & 2-hour & 600.14 & 340.06 & 13.98 & 0.93 \\
 & 3-hour & 660.65 & 372.04 & 15.18 & 0.91 \\
 & 4-hour & 708.24 & 396.25 & 16.23 & 0.90 \\
 & 5-hour & 738.01 & 411.27 & 17.10 & 0.89 \\
 & 6-hour & 760.32 & 423.74 & 17.91 & 0.88 \\
 & Overall & 627.60 & 354.72 & 14.67 & 0.92 \\
\midrule

CNN--LSTM & 1-hour & 541.02 & 301.83 & 12.61 & 0.94 \\
 & 2-hour & 603.64 & 346.53 & 14.10 & 0.93 \\
 & 3-hour & 661.03 & 377.12 & 15.35 & 0.91 \\
 & 4-hour & 706.17 & 401.15 & 16.32 & 0.90 \\
 & 5-hour & 731.89 & 415.12 & 17.11 & 0.89 \\
 & 6-hour & 753.62 & 426.54 & 17.86 & 0.88 \\
 & Overall & 626.53 & 357.59 & 14.72 & 0.92 \\
\midrule

GCN & 1-hour & 631.79 & 384.45 & 16.48 & 0.92 \\
 & 2-hour & 691.51 & 429.72 & 17.82 & 0.91 \\
 & 3-hour & 744.90 & 462.17 & 18.96 & 0.89 \\
 & 4-hour & 783.01 & 487.10 & 19.86 & 0.88 \\
 & 5-hour & 805.12 & 496.76 & 20.49 & 0.87 \\
 & 6-hour & 827.87 & 509.45 & 21.73 & 0.86 \\
 & Overall & 713.37 & 440.32 & 18.25 & 0.90 \\
\midrule

SVR & 1-hour & 665.10 & 449.35 & 20.48 & 0.91 \\
 & 2-hour & 762.21 & 510.90 & 21.77 & 0.89 \\
 & 3-hour & 834.45 & 559.10 & 22.80 & 0.86 \\
 & 4-hour & 877.51 & 587.21 & 23.65 & 0.84 \\
 & 5-hour & 901.36 & 600.37 & 24.45 & 0.83 \\
 & 6-hour & 916.22 & 602.64 & 25.30 & 0.83 \\
 & Overall & 780.66 & 516.54 & 21.65 & 0.88 \\
\midrule

GCN--LSTM & 1-hour & 295.52 & 158.03 & 6.45 & 0.98 \\
 & 2-hour & 400.46 & 207.14 & 8.26 & 0.97 \\
 & 3-hour & 483.98 & 242.88 & 9.64 & 0.95 \\
 & 4-hour & 543.44 & 267.39 & 10.70 & 0.94 \\
 & 5-hour & 578.53 & 285.67 & 11.71 & 0.93 \\
 & 6-hour & 610.84 & 306.17 & 12.82 & 0.92 \\
 & Overall & 447.01 & 230.35 & 9.34 & 0.96 \\
\midrule

\textbf{GT--LSTM (proposed)}
& 1-hour & \textbf{291.74} & \textbf{154.72} & \textbf{6.41} & \textbf{0.98} \\
& 2-hour & \textbf{394.63} & \textbf{201.87} & \textbf{8.00} & \textbf{0.97} \\
& 3-hour & \textbf{473.40} & \textbf{233.97} & \textbf{9.28} & \textbf{0.96} \\
& 4-hour & \textbf{531.37} & \textbf{258.00} & \textbf{10.25} & \textbf{0.94} \\
& 5-hour & \textbf{566.03} & \textbf{275.96} & \textbf{11.24} & \textbf{0.93} \\
& 6-hour & \textbf{594.59} & \textbf{292.62} & \textbf{12.37} & \textbf{0.93} \\
& \textbf{Overall}
& \textbf{435.11}
& \textbf{221.27}
& \textbf{8.90}
& \textbf{0.96} \\

\bottomrule
\end{tabular}%
}

\endgroup

\vspace{4pt}

\begin{minipage}{0.95\textwidth}
\footnotesize
\textit{Note.} Minimum flow = 39; maximum flow = 15,873;
mean flow = 3,649.40; median flow = 3,757.
\end{minipage}

\end{table}

\begin{table}[htbp]
    \centering

    \caption{Highway-specific traffic-flow prediction performance}
    \label{tab:roadway_specific_metrics}

    \renewcommand{\arraystretch}{1.05}

    \resizebox{\textwidth}{!}{%
    \begin{tabular}{lcccccccccccc}
        \toprule

        & \multicolumn{4}{c}{\textbf{I-101}}
        & \multicolumn{4}{c}{\textbf{I-405}}
        & \multicolumn{4}{c}{\textbf{I-210}} \\

        \cmidrule(lr){2-5}
        \cmidrule(lr){6-9}
        \cmidrule(lr){10-13}

        \textbf{Horizon}
        & \textbf{RMSE} & \textbf{MAE} & \textbf{SMAPE} & \textbf{$R^2$}
        & \textbf{RMSE} & \textbf{MAE} & \textbf{SMAPE} & \textbf{$R^2$}
        & \textbf{RMSE} & \textbf{MAE} & \textbf{SMAPE} & \textbf{$R^2$} \\

        \midrule

        1-hour
        & 393.2 & 196.7 & 5.76 & 0.97
        & 356.8 & 203.0 & 5.38 & 0.97
        & 396.2 & 244.2 & 12.05 & 0.95 \\

        2-hour
        & 537.4 & 265.3 & 7.96 & 0.94
        & 476.0 & 255.1 & 6.57 & 0.95
        & 548.4 & 340.4 & 15.79 & 0.91 \\

        3-hour
        & 659.7 & 318.6 & 9.90 & 0.91
        & 556.0 & 289.4 & 7.44 & 0.93
        & 663.7 & 412.7 & 18.96 & 0.87 \\

        4-hour
        & 746.2 & 356.6 & 11.38 & 0.88
        & 605.8 & 313.1 & 8.09 & 0.92
        & 750.9 & 468.5 & 21.33 & 0.83 \\

        5-hour
        & 797.5 & 385.7 & 12.64 & 0.86
        & 607.8 & 319.9 & 8.51 & 0.92
        & 807.7 & 504.5 & 23.31 & 0.80 \\

        6-hour
        & 839.4 & 414.1 & 13.96 & 0.85
        & 606.5 & 321.6 & 8.86 & 0.92
        & 861.4 & 545.1 & 25.43 & 0.76 \\

        \midrule

        \textbf{Overall}
        & \textbf{680.2} & \textbf{322.8} & \textbf{10.27} & \textbf{0.90}
        & \textbf{542.7} & \textbf{283.7} & \textbf{7.47} & \textbf{0.94}
        & \textbf{690.0} & \textbf{419.2} & \textbf{19.48} & \textbf{0.86} \\

        \bottomrule
    \end{tabular}%
    }

    \vspace{4pt}

    \begin{minipage}{0.98\textwidth}
    \footnotesize
    \textit{Note.}
    For I-101, minimum = 60.0, maximum = 9,921.0,
    mean = 4,247.39, and median = 4,565.0.
    For I-405, minimum = 60.0, maximum = 11,021.0,
    mean = 4,528.56, and median = 5,123.0.
    For I-210, minimum = 59.0, maximum = 7,792.0,
    mean = 2,725.16, and median = 2,400.0.
    SMAPE values are expressed as percentages.
    \end{minipage}

\end{table}

\begin{figure}[htbp]
\centering

\subfloat[
I-101 detector at Station 716391
\label{fig:i101}
]{%
    \includegraphics[
        width=0.75\textwidth
    ]{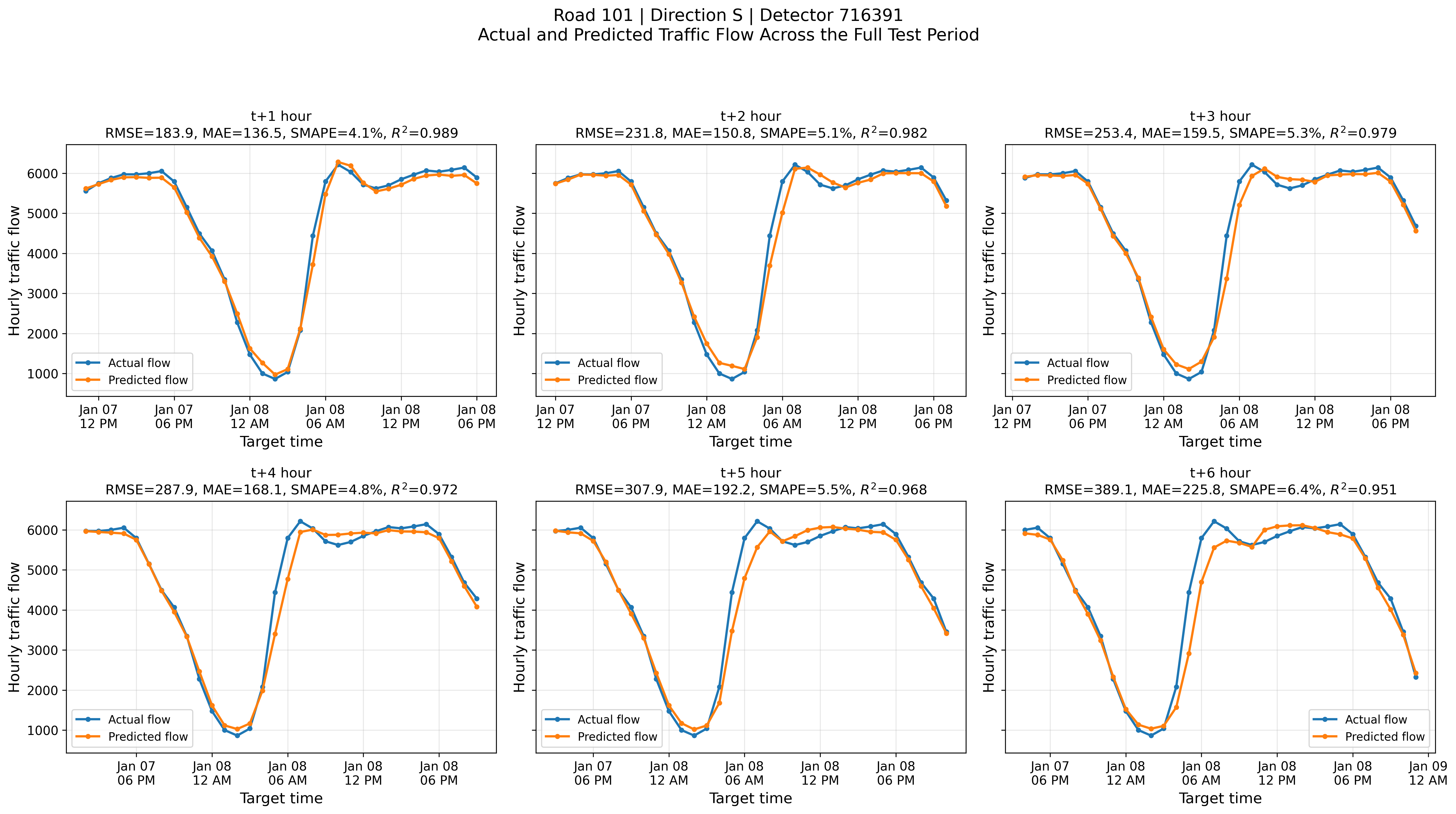}
}
\hfill
\subfloat[
I-405 detector at Station 718238
\label{fig:i405}
]{%
    \includegraphics[
        width=0.75\textwidth
    ]{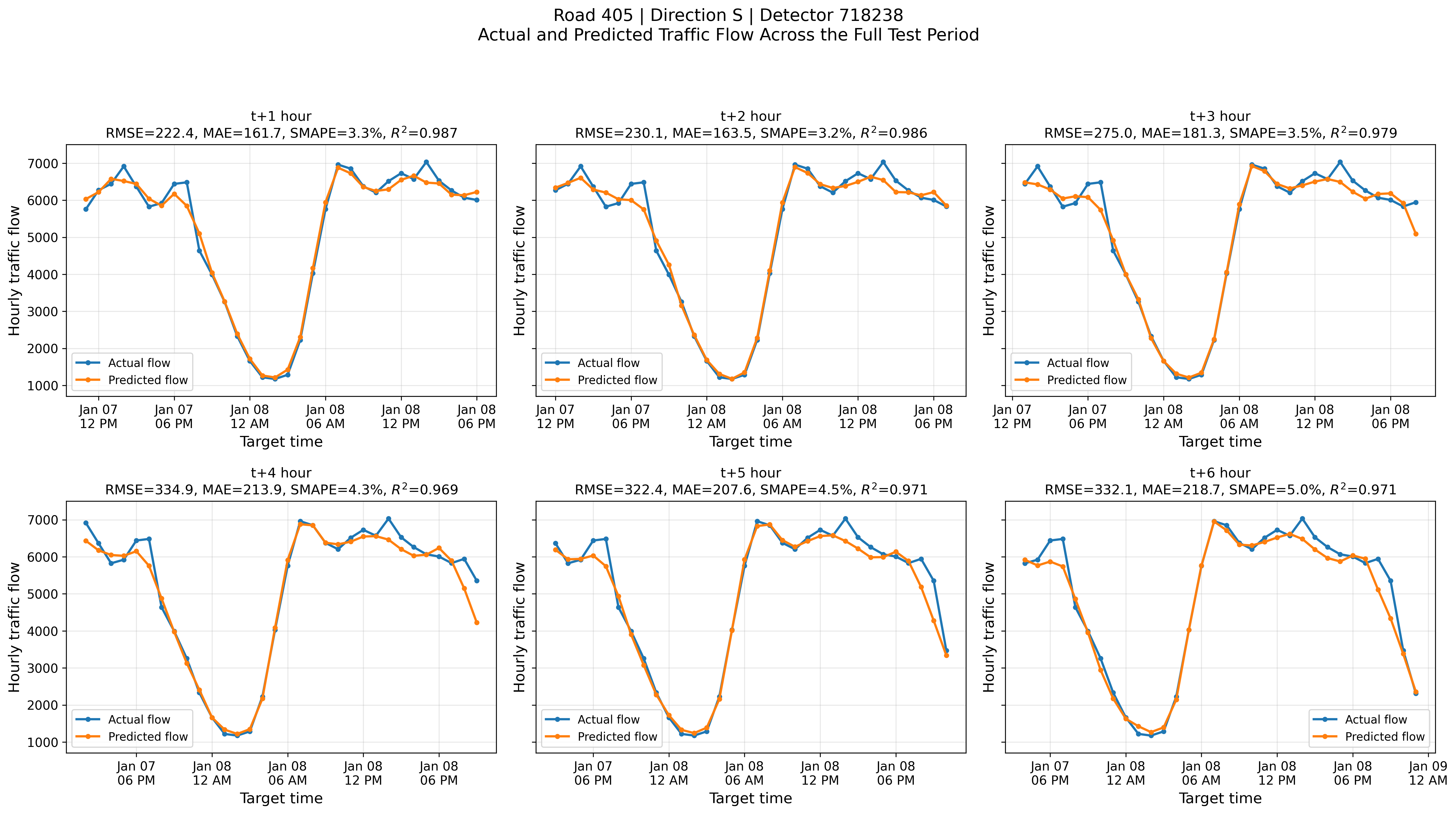}
}

\par\medskip

\subfloat[
I-210 detector at Station 717659
\label{fig:i210}
]{%
    \includegraphics[
        width=0.75\textwidth
    ]{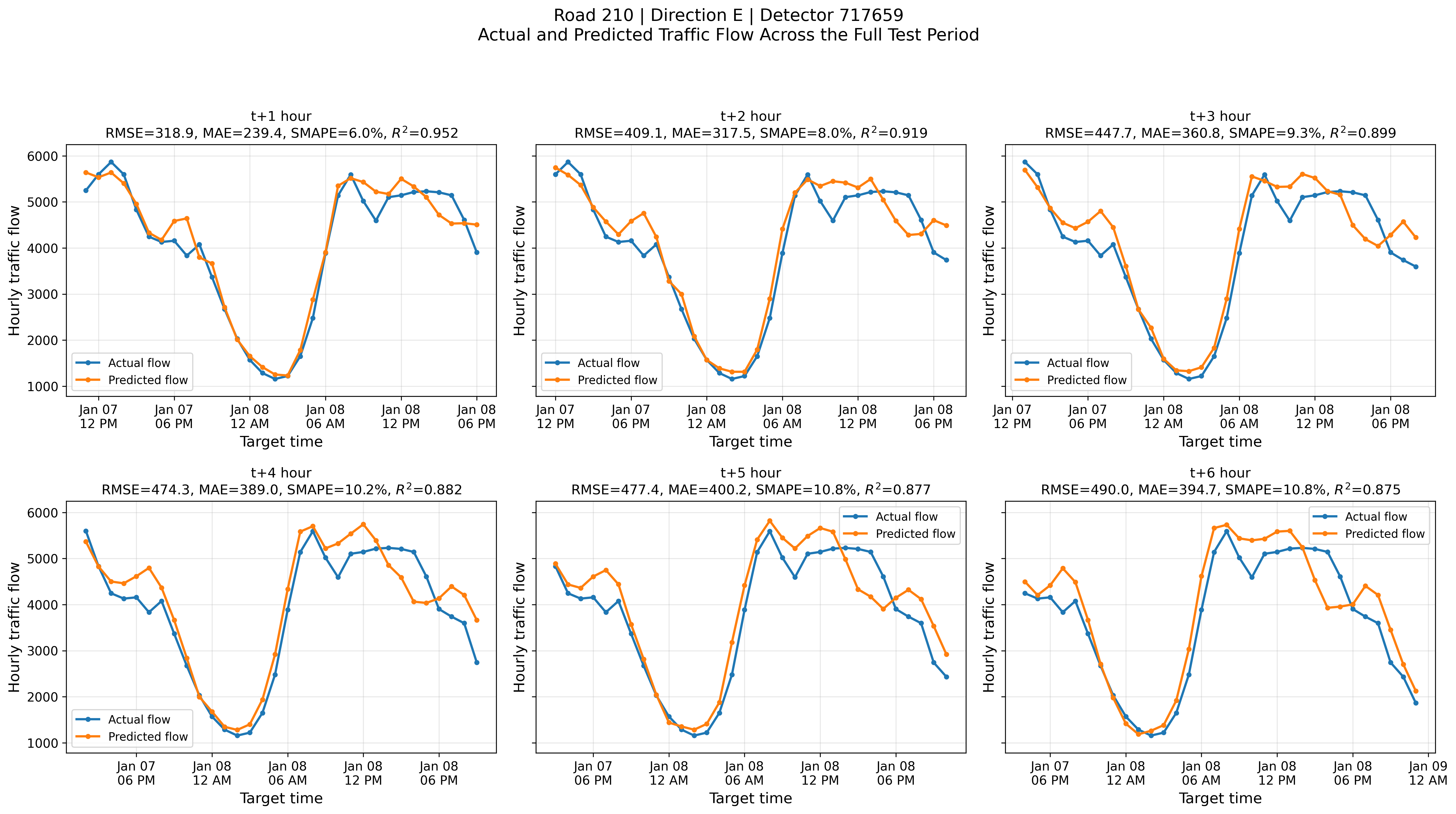}
}

\caption{Observed and predicted traffic flow at representative detectors on I-101, I-405, and I-210 evacuation corridors.}
\label{fig:line_plot}

\end{figure}

Table~\ref{tab:roadway_specific_metrics} reports prediction performance for I-101, I-405, and I-210. These corridors were examined because they are major highways near the Palisades and Eaton fire-affected areas and may carry substantial evacuation-related traffic. The results show that model performance varied across the three corridors. Among the three highways, I-405 produced the highest overall prediction accuracy. The model achieved an overall RMSE of 542.7, an MAE of 283.7, a SMAPE of 7.47\%, and an $R^2$ of 0.94. At the six-hour horizon, the $R^2$ remained 0.92 and the SMAPE remained below 9\%. It indicates that the temporal evolution of traffic on I-405 was comparatively stable and predictable over the six-hour forecasting period. The representative I-405 detector, Station 718238, also exhibited close agreement between the observed and predicted traffic-flow profiles (Figure~\ref{fig:line_plot}). The model captured the evening decline, the overnight low-flow period, and the rapid morning increase. Prediction performance remained particularly stable across the six horizons. The $R^2$ decreased only from 0.987 at the one-hour horizon to 0.971 at the six-hour horizon, while SMAPE increased from 3.3\% to 5.0\%. The longer-horizon predictions slightly underestimated some traffic values near the end of the test period, but the general temporal pattern remained accurately represented. The detector-level results are consistent with the overall highway-level performance observed for I-405.

The model also demonstrated strong performance on I-101, although the errors increased more rapidly with the forecast horizon. At the one-hour horizon, the model achieved an $R^2$ of 0.97 and a SMAPE of 5.76\%. At the six-hour horizon, the RMSE increased from 393.2 to 839.4, while the $R^2$ decreased from 0.97 to 0.85. Thus, the model accurately captured short-term traffic conditions on I-101, but uncertainty increased for longer prediction horizons. At a representative station 716391 of I-101, the model closely followed the observed traffic-flow profile across all six prediction horizons (Figure~\ref{fig:line_plot}). The detector experienced a pronounced decline from approximately 6,000 vehicles to less than 1,000 vehicles during the overnight period, followed by a rapid recovery during the morning. The GT--LSTM model reproduced both the magnitude and timing of this decline and subsequent increase. The deviations at the longer horizons occurred during the rapid morning recovery, where the predicted increase was slightly delayed or smoothed. This pattern indicates that abrupt transitions were more difficult to predict than periods of stable flow.

I-210 presented the most challenging prediction environment. The model achieved an overall RMSE of 690.0, an MAE of 419.2, a SMAPE of 19.48\%, and an $R^2$ of 0.86. The one-hour prediction remained strong, with an $R^2$ of 0.95. However, the $R^2$ decreased to 0.76 at the six-hour horizon, while the SMAPE increased from 12.05\% to 25.43\%. Although the overall RMSE for I-210 was similar to that for I-101, its substantially higher SMAPE indicates larger errors relative to the magnitude of the observed traffic flow. This difference is partly associated with the lower average traffic flow on I-210. The mean flow on I-210 was 2,725.16 vehicles, compared with 4,247.39 vehicles on I-101 and 4,528.56 vehicles on I-405. Therefore, an absolute prediction error of a similar magnitude produces a larger percentage error on I-210. At a representative station 717659 of I-210, the model reproduced the general overnight decline and morning recovery (Figure~\ref{fig:line_plot}). Larger deviations occurred after the morning recovery and during the later portion of the test period. The model tended to smooth some of the local fluctuations and occasionally overestimated or underestimated the magnitude of the observed traffic changes. Overall, I-210 contained less predictable temporal behavior than I-101 and I-405. The overall results suggest that the traffic pattern on I-405 was more temporally persistent, while I-210 exhibited more irregular changes that were difficult to anticipate several hours in advance.

\begin{figure}[htbp]
\centering


\subfloat[
January 8, 2025, 4:00 a.m.
\label{fig:network_propagation_t1}
]{%
    \includegraphics[
        width=0.48\textwidth
    ]{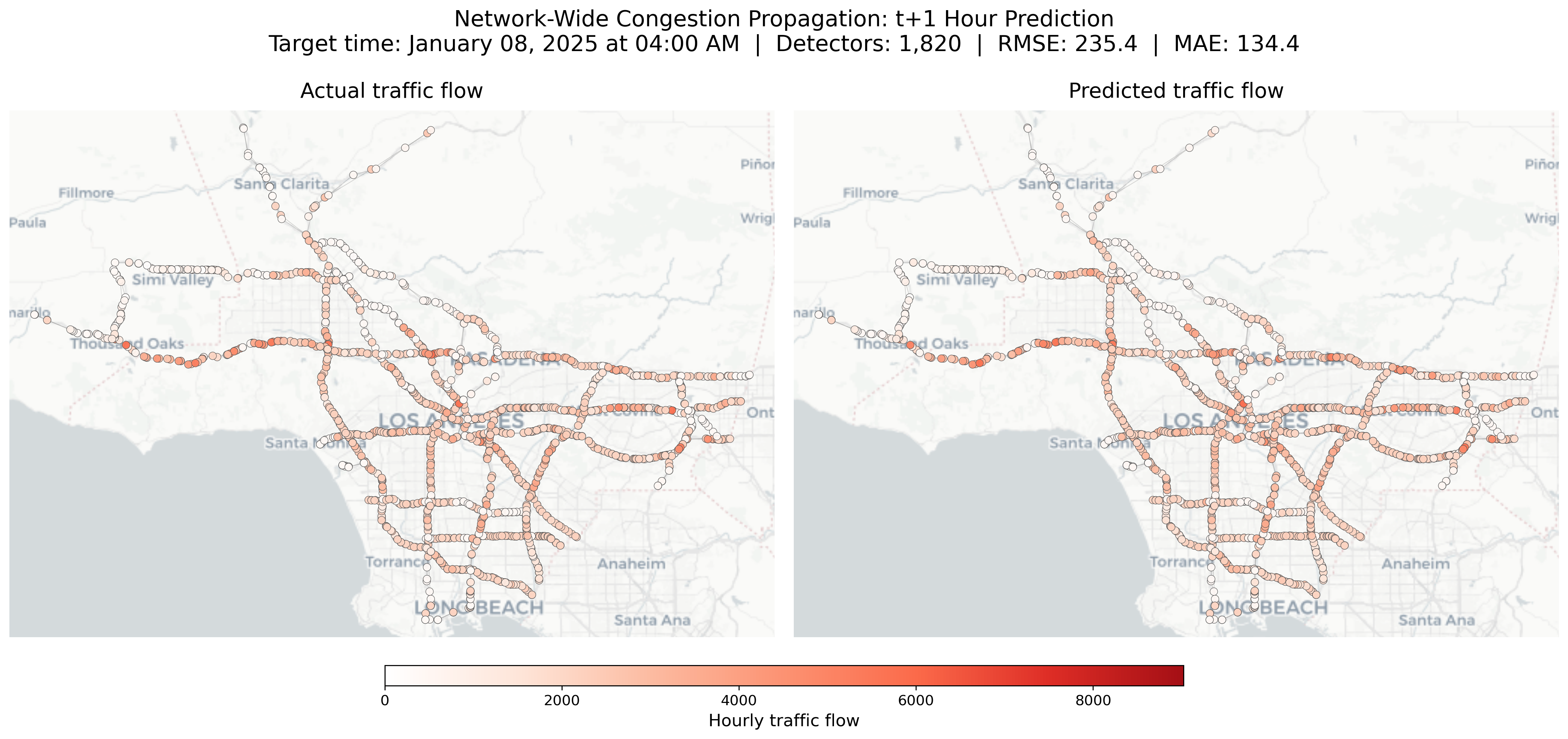}
}
\hfill
\subfloat[
January 8, 2025, 5:00 a.m.
\label{fig:network_propagation_t2}
]{%
    \includegraphics[
        width=0.48\textwidth
    ]{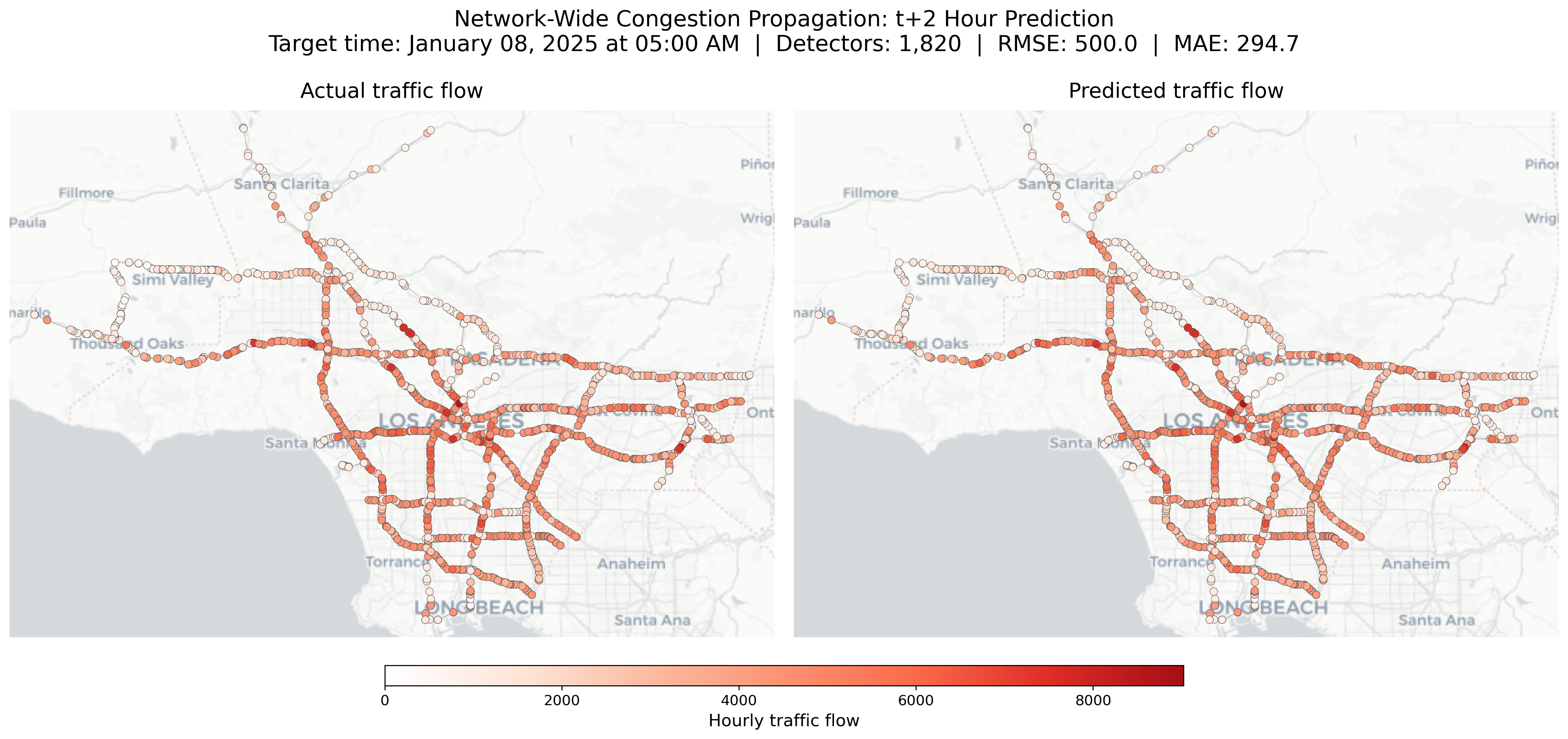}
}

\par\medskip


\subfloat[
January 8, 2025, 6:00 a.m.
\label{fig:network_propagation_t3}
]{%
    \includegraphics[
        width=0.48\textwidth
    ]{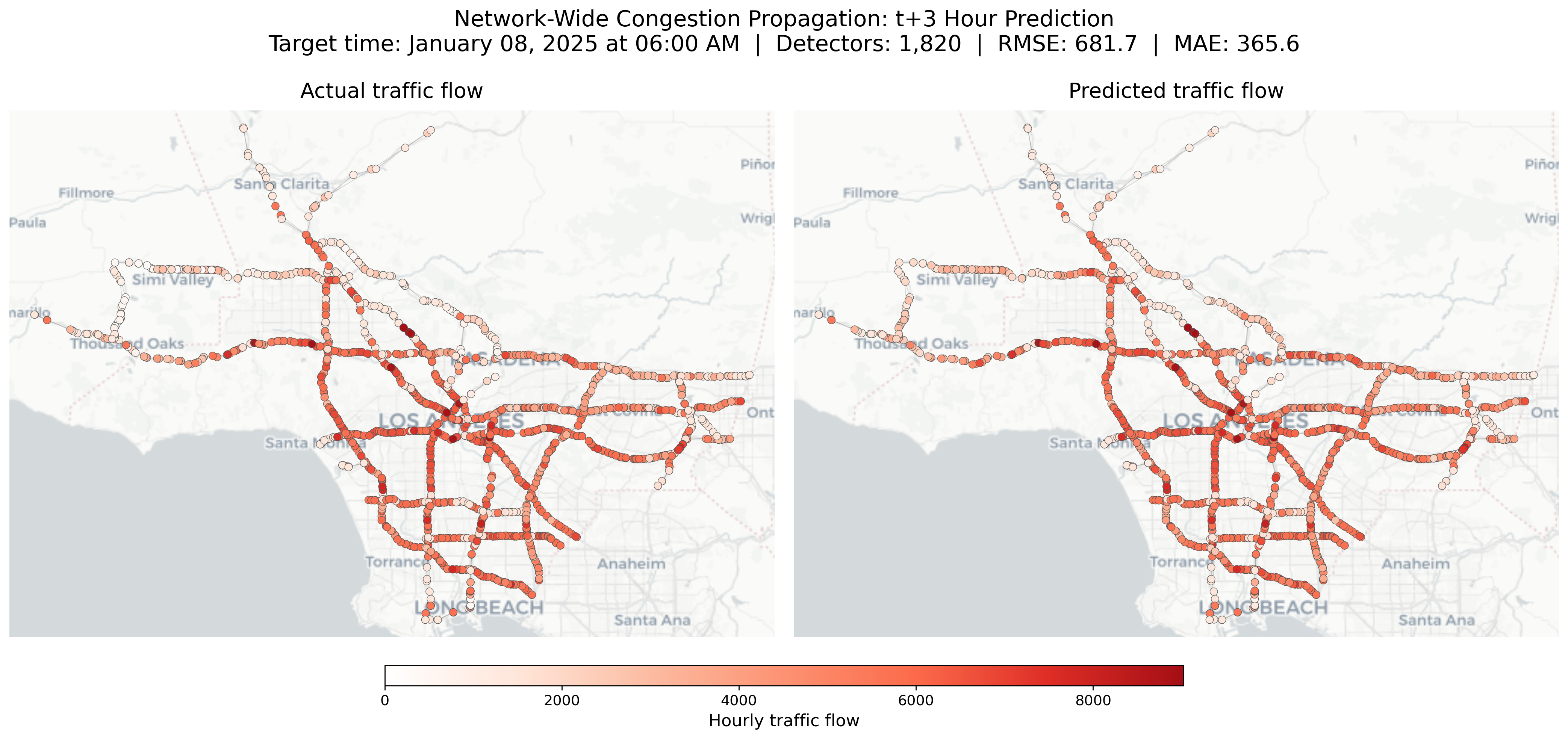}
}
\hfill
\subfloat[
January 8, 2025, 7:00 a.m.
\label{fig:network_propagation_t4}
]{%
    \includegraphics[
        width=0.48\textwidth
    ]{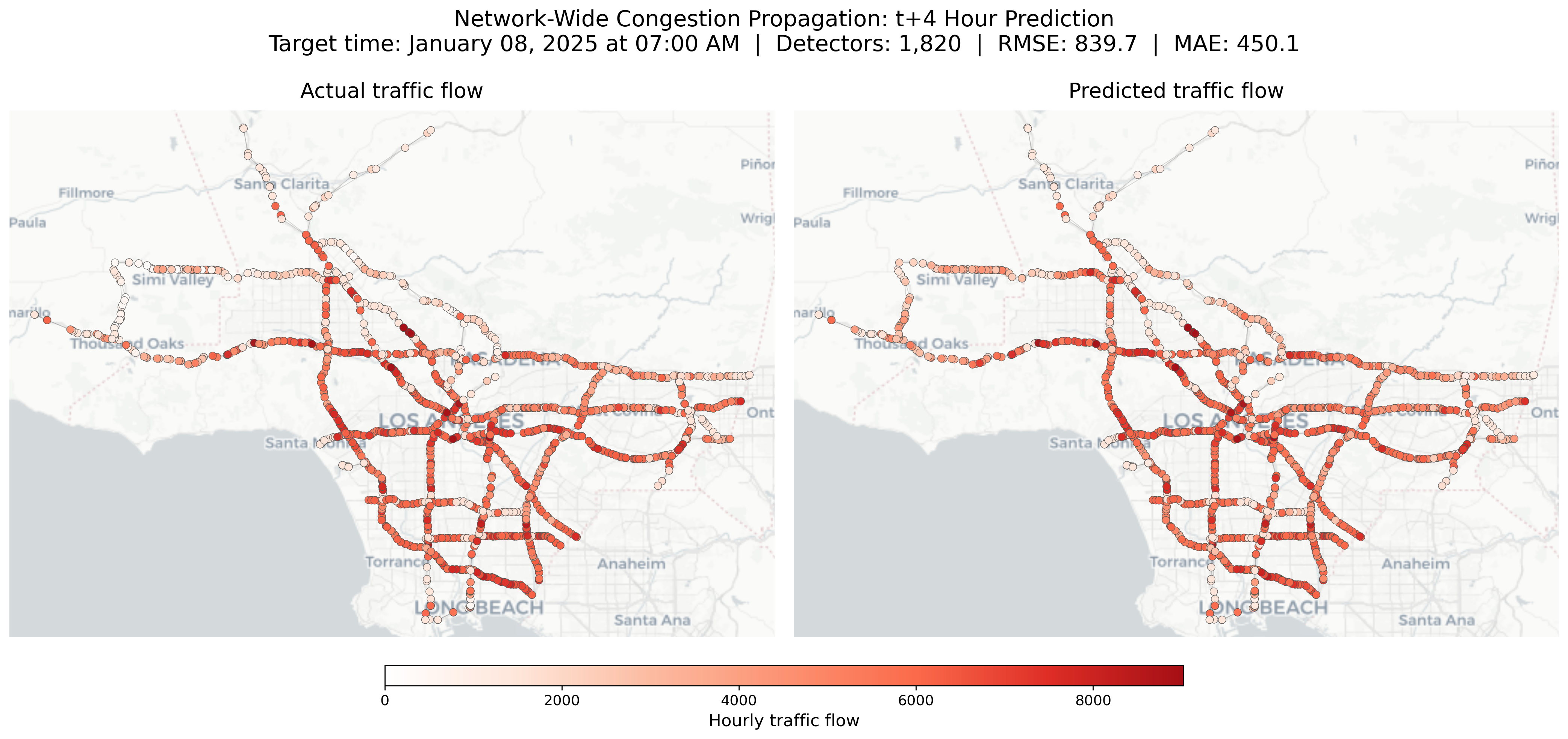}
}

\par\medskip


\subfloat[
January 8, 2025, 8:00 a.m.
\label{fig:network_propagation_t5}
]{%
    \includegraphics[
        width=0.48\textwidth
    ]{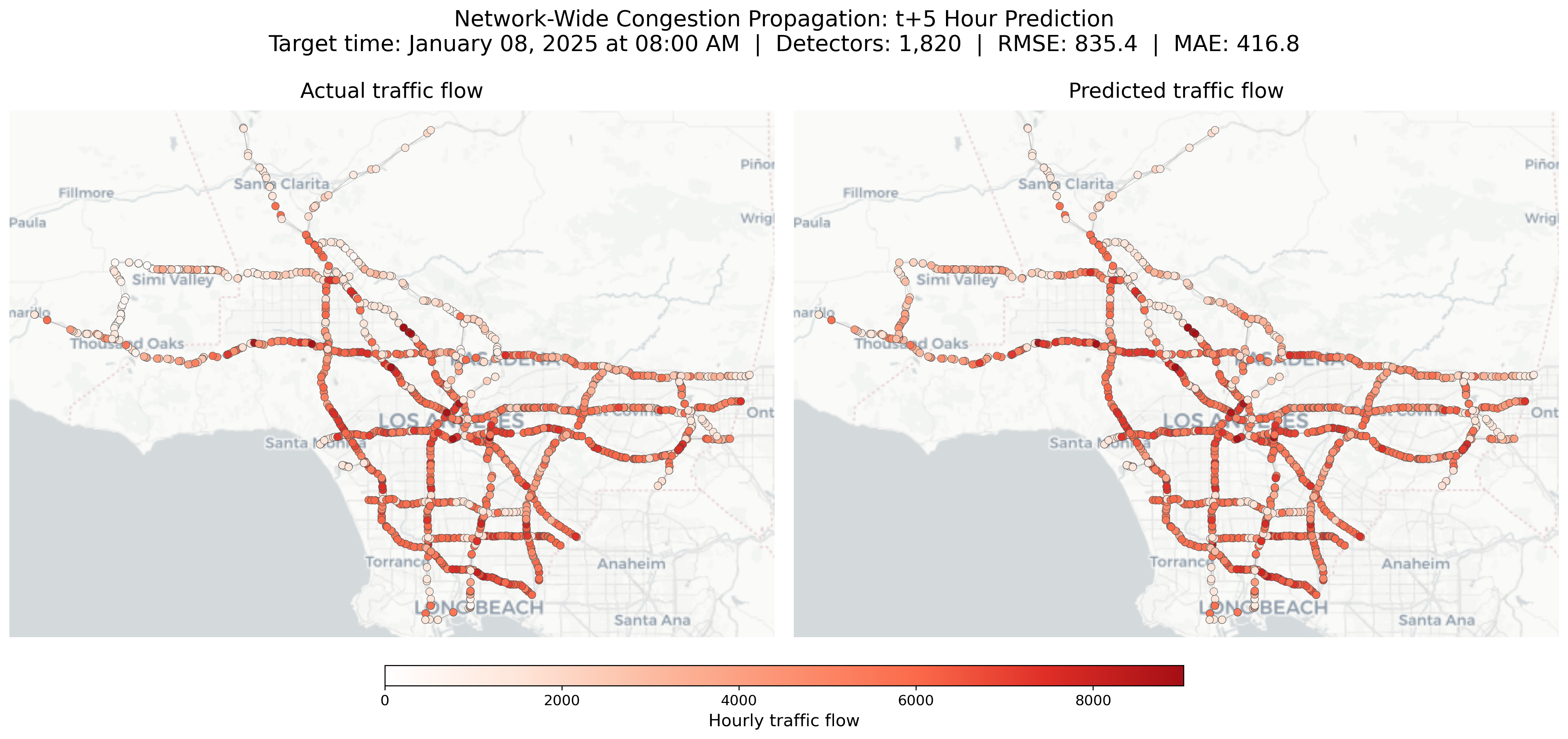}
}
\hfill
\subfloat[
January 8, 2025, 9:00 a.m.
\label{fig:network_propagation_t6}
]{%
    \includegraphics[
        width=0.48\textwidth
    ]{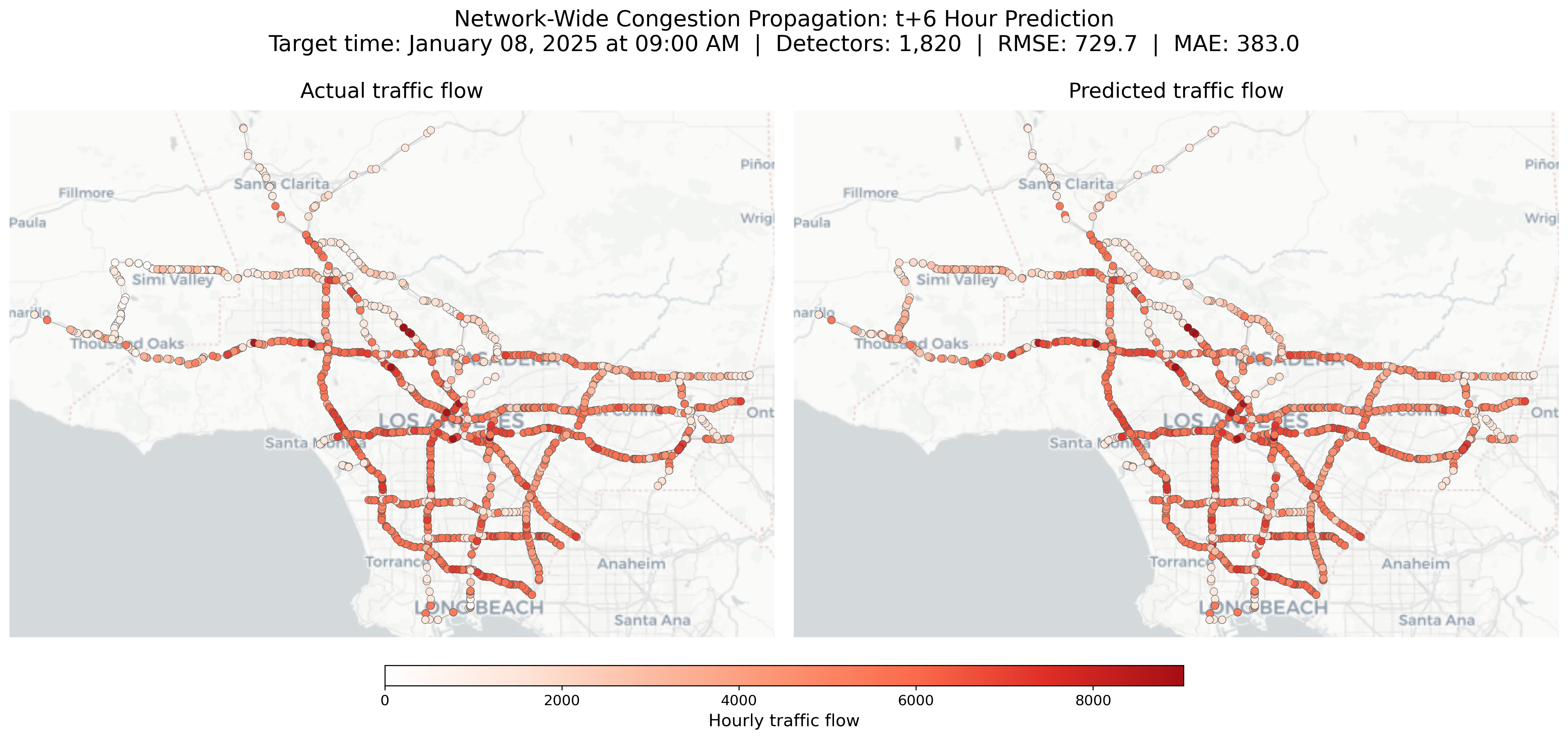}
}

\caption{Network-wide spatial distributions of observed and predicted traffic flow across the Caltrans District 7 network for predictions generated at 3:00 a.m. on January 8, 2025.}
\label{fig:network_congestion_propagation}

\end{figure}

Figure~\ref{fig:network_congestion_propagation} compares the spatial distributions of observed and predicted traffic flow across the 1,820 detector stations for the one- through six-hour prediction horizons. The panels illustrate how the GT--LSTM model reproduced changes in traffic-flow intensity across the District 7 network during the transition from overnight conditions to the morning period on January 8, 2025. At 4:00 a.m., corresponding to the one-hour prediction horizon, relatively low traffic flow was observed across most of the network. At 5:00 a.m., traffic flow began to increase along several major highways. The model captured the emerging increases across the central, western, and southern portions of the network. The RMSE and MAE increased to 500.0 and 294.7, respectively. The higher errors reflect both the longer forecast horizon and the transition from stable overnight traffic to rapidly changing morning conditions. The 6:00 a.m. and 7:00 a.m. high-flow conditions became more spatially extensive along the most of the corridors of the network. The predicted maps reproduced the general locations and spatial continuity of these high-flow patterns. However, some differences are visible in the magnitude of traffic flow at individual detector stations. The RMSE values were 681.7 and 839.7 for the three- and four-hour horizons, respectively, while the corresponding MAE values were 365.6 and 450.1. The higher errors at these times indicate that rapid, spatially distributed morning traffic increases were more difficult to predict than the preceding low-flow conditions. At 8:00 a.m. and 9:00 a.m., high traffic flow remained widespread across the network. The GT--LSTM predictions continued to reproduce the major high-flow corridors and the overall spatial distribution of traffic demand. Overall, the spatial comparisons show that the GT--LSTM model captured the development and propagation of network-wide traffic-flow patterns during the wildfire evacuation period. The predicted maps identify the major high-flow corridors and reproduce the transition from low overnight flow to widespread morning traffic activity. 

\section{Ablation Experiments}
\begin{figure}[htbp]
    \centering
    \includegraphics[
        width=\textwidth
    ]{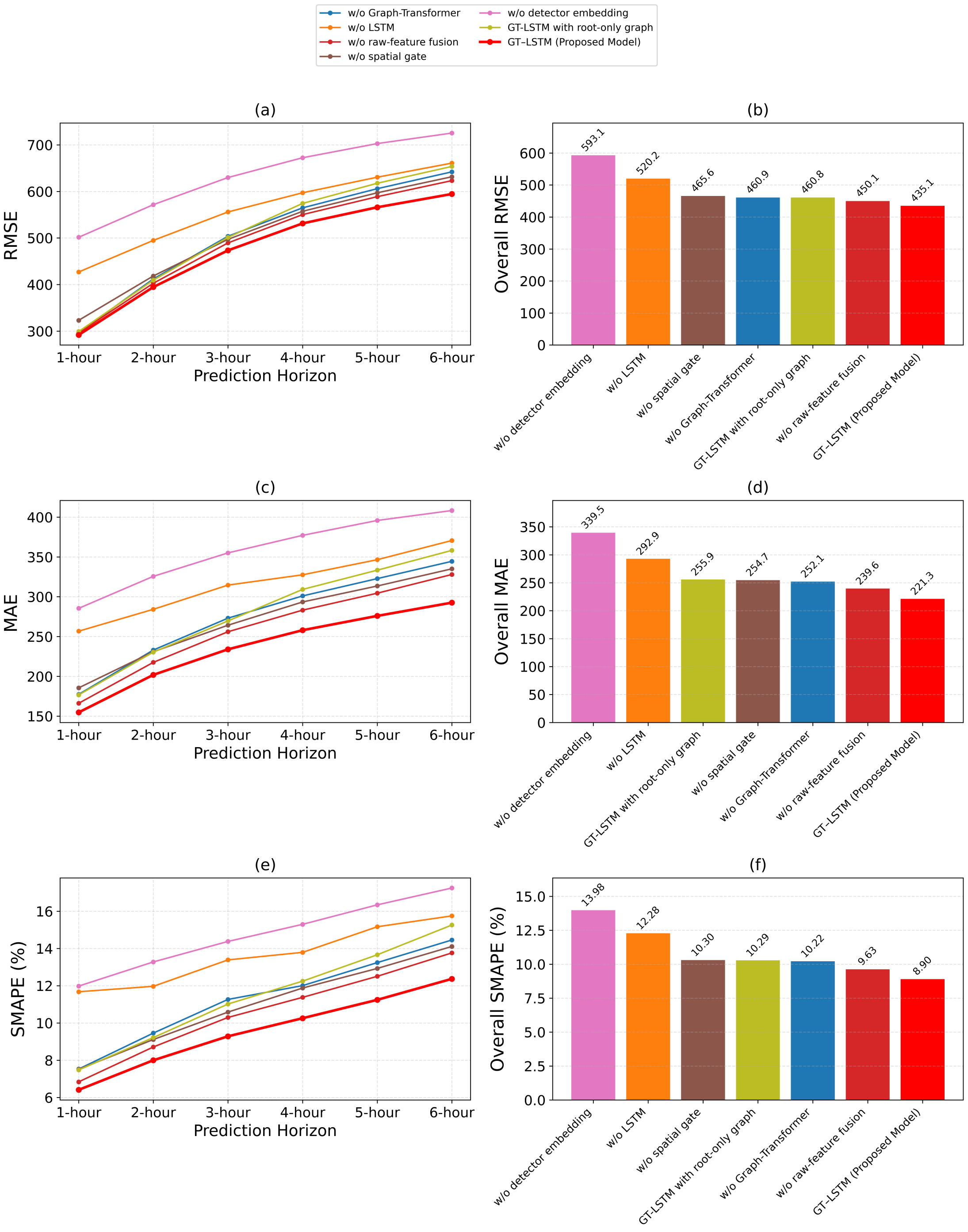}
    \caption{Ablation analysis of the proposed GT--LSTM model.}
    \label{fig:wildfire_ablation_comparison}
\end{figure}

An ablation analysis was conducted to evaluate the contribution of the major components of the proposed GT--LSTM framework. Figure~\ref{fig:wildfire_ablation_comparison} presents the prediction performance after removing or modifying one component at a time. For each ablation experiment, all model hyperparameters were kept the same as those used in the full model so that the change in performance could be attributed to the specific component being evaluated. Among the tested components, the detector embedding had the largest influence on prediction performance. Removing the detector embedding increased the overall RMSE from 435.11 to 593.10, the MAE from 221.27 to 339.54, and the SMAPE from 8.90\% to 13.98\%. This result shows that persistent detector-specific characteristics are important for representing location-specific traffic patterns that may not be fully captured by the observed model features. Removing the LSTM also caused a substantial reduction in performance, increasing the overall RMSE to 520.24 and the SMAPE to 12.28\%. The degradation became more noticeable at longer prediction horizons, highlighting the importance of modeling temporal dependencies in the six-hour input sequence.

The graph-transformer, spatial gate, and direct raw-feature fusion also contributed to the prediction performance. Removing the graph-transformer branch increased the overall RMSE to 460.94 and the MAE to 252.12. A similar result was observed for the root-only graph configuration, which produced an overall RMSE of 460.83. Since the root-only configuration retains the root transformation but removes information exchange between connected detectors, this result indicates that incorporating information from upstream detectors contributes to the spatial learning capability of the model. Removing the spatial gate also increased the overall RMSE to 465.62, MAE to 254.72, and SMAPE to 10.30\%, showing the benefit of adaptively controlling the contribution of spatially aggregated information. Similarly, removing the direct raw-feature fusion increased the overall RMSE to 450.05 and the SMAPE to 9.63\%, indicating that providing the LSTM with direct access to the original standardized features is also beneficial.

The full GT--LSTM achieved the lowest prediction errors across the tested configurations. The ablation results show that the model performance is not driven by a single component. The detector embedding and LSTM provided the largest improvements, while the graph-transformer branch, spatial gate, and direct raw-feature fusion provided additional performance gains. These results support the combined use of location-specific, spatial, and temporal information for network-wide wildfire evacuation traffic prediction.

\section{Conclusions}
\label{sec:conclusions}
This study developed a Graph Transformer--Long Short-Term Memory (GT--LSTM) framework for network-wide wildfire evacuation traffic prediction. The model achieved strong performance across all prediction horizons. The GT--LSTM outperformed the LSTM, CNN, CNN--LSTM, GCN, support vector regression, and GCN--LSTM baseline models across the reported error measures. Ablation experiments further confirmed that each major model component contributed to improving the prediction performance, with the detector embedding and LSTM showing the largest effects. The corridor-level and detector-level results further showed that the model captured major temporal changes and spatial traffic-flow patterns on evacuation-related highways. These findings demonstrate that spatial and temporal patterns learned from routine detector observations can support network-wide traffic forecasting during an unseen wildfire event, even when detailed wildfire-related information is unavailable. 

Emergency management officials can use the real-time forecasts generated by the model to anticipate traffic conditions across different parts of the network before severe congestion develops. This information can support proactive decisions related to evacuation-route management, traffic-control deployment, traveler information, and the allocation of emergency response resources. By identifying locations and time periods where traffic demand is expected to increase in real-time, agencies can intervene earlier and improve the overall efficiency and safety of evacuation operations. 

This study has some limitations. First, the model was evaluated using the Palisades and Eaton fires, which occurred on the same day within a single regional transportation network. Additional wildfire events and roadway networks should be examined before the findings can be generalized to other geographic and operational conditions. Second, the model did not include wildfire-specific variables, such as evacuation orders, fire spread, evacuation-zone boundaries, roadway closures, incidents, smoke conditions, population displacement, or emergency traffic-control actions. Although this study demonstrates the value of routinely available traffic data, the omission of these variables may limit prediction accuracy when wildfire conditions cause abrupt changes that are not evident in the recent traffic history. Third, the roadway graph and its edge attributes remained fixed during the prediction period and therefore did not represent time-varying roadway closures, detector failures, or changes in network accessibility. Future research should evaluate the model across multiple wildfire events, incorporate real-time hazard and transportation-system information, and assess how the resulting forecasts can support evacuation route guidance and emergency traffic-management decisions.

\section*{Author Contributions Statement (CRediT)}
\textbf{M. Rafi:} Conceptualization, Data curation, Formal analysis, Investigation, Methodology, Validation, Visualization, Writing – original draft.  
\textbf{S. Hasan:} Conceptualization, Methodology, Resources, Supervision, Project administration, Validation, Writing – review and editing. Both authors reviewed and approved the final manuscript.

\section*{Data Availability Statement}
The raw detector data used in this study were collected from Performance Measurement System (PeMS) of California Department of Transportation. Raw data can be directly downloaded from this link: \nolinkurl{https://pems.dot.ca.gov/?dnode=Clearinghouse&type=station_hour&district_id=7}

\section*{Declaration of Conflicting Interests}
The authors declare that they have no competing financial interests or personal relationships that could have influenced the work reported in this study.

\section*{Declaration of Generative AI}
During the preparation of this manuscript, the authors used ChatGPT for assistance in language refinement and LaTeX formatting. After using this tool/service, the authors reviewed and edited the content as needed and take full responsibility for the content of the manuscript.

\section*{Declaration of Funding}
No funding was received for this research.

\pagebreak 
\bibliography{all.bib}

@article{RadeloffEtAl2018,
author = {Radeloff, Volker C. and Helmers, David P. and Kramer, H. Anu and Mockrin, Miranda H. and Alexandre, Patricia M. and Bar-Massada, Avi and Butsic, Van and Hawbaker, Todd J. and Martinuzzi, Sebasti{'a}n and Syphard, Alexandra D. and Stewart, Susan I.},
title = {Rapid Growth of the {US} Wildland--Urban Interface Raises Wildfire Risk},
journal = {Proceedings of the National Academy of Sciences},
year = {2018},
volume = {115},
number = {13},
pages = {3314--3319},
doi = {10.1073/pnas.1718850115}
}

@article{CovaJohnson2002,
author = {Cova, Thomas J. and Johnson, Justin P.},
title = {Microsimulation of Neighborhood Evacuations in the Urban--Wildland Interface},
journal = {Environment and Planning A},
year = {2002},
volume = {34},
number = {12},
pages = {2211--2229},
doi = {10.1068/a34251}
}

@article{IntiniEtAl2019,
author = {Intini, Paolo and Ronchi, Enrico and Gwynne, Steven M. V. and Pel, Adam},
title = {Traffic Modeling for Wildland--Urban Interface Fire Evacuation},
journal = {Journal of Transportation Engineering, Part A: Systems},
year = {2019},
volume = {145},
number = {3},
pages = {04019002},
doi = {10.1061/JTEPBS.0000221}
}

@article{Kuligowski2021,
author = {Kuligowski, Erica D.},
title = {Evacuation Decision-Making and Behavior in Wildfires: Past Research, Current Challenges and a Future Research Agenda},
journal = {Fire Safety Journal},
year = {2021},
volume = {120},
pages = {103129},
doi = {10.1016/j.firesaf.2020.103129}
}

@article{WahlqvistEtAl2021,
author = {Wahlqvist, Jonathan and Ronchi, Enrico and Gwynne, Steven M. V. and Kinateder, Max and Rein, Guillermo and Mitchell, Harry and B{'e}nichou, Noureddine and Ma, Chunyun and Kimball, Amanda and Kuligowski, Erica D.},
title = {The Simulation of Wildland--Urban Interface Fire Evacuation: The {WUI-NITY} Platform},
journal = {Safety Science},
year = {2021},
volume = {136},
pages = {105145},
doi = {10.1016/j.ssci.2020.105145}
}

@article{ZhaoEtAl2022,
author = {Zhao, Xilei and Xu, Yiming and Lovreglio, Ruggiero and Kuligowski, Erica D. and Nilsson, Daniel and Cova, Thomas J. and Wu, Alex and Yan, Xiang},
title = {Estimating Wildfire Evacuation Decision and Departure Timing Using Large-Scale {GPS} Data},
journal = {Transportation Research Part D: Transport and Environment},
year = {2022},
volume = {107},
pages = {103277},
doi = {10.1016/j.trd.2022.103277}
}

@article{GrajduraEtAl2022,
author = {Grajdura, Sarah and Borjigin, Sachraa and Niemeier, Deb},
title = {Fast-Moving Dire Wildfire Evacuation Simulation},
journal = {Transportation Research Part D: Transport and Environment},
year = {2022},
volume = {104},
pages = {103190},
doi = {10.1016/j.trd.2022.103190}
}

@article{RohaertEtAl2023Kincade,
author = {Rohaert, Arthur and Kuligowski, Erica D. and Ardinge, Adam and Wahlqvist, Jonathan and Gwynne, Steven M. V. and Kimball, Amanda and B{'e}nichou, Noureddine and Ronchi, Enrico},
title = {Traffic Dynamics during the 2019 {Kincade} Wildfire Evacuation},
journal = {Transportation Research Part D: Transport and Environment},
year = {2023},
volume = {116},
pages = {103610},
doi = {10.1016/j.trd.2023.103610}
}

@article{RohaertEtAl2023Glass,
author = {Rohaert, Arthur and Janfeshanaraghi, Nima and Kuligowski, Erica D. and Ronchi, Enrico},
title = {The Analysis of Traffic Data of Wildfire Evacuation: The Case Study of the 2020 {Glass} Fire},
journal = {Fire Safety Journal},
year = {2023},
volume = {141},
pages = {103909},
doi = {10.1016/j.firesaf.2023.103909}
}

@article{HouEtAl2023,
author = {Hou, Zenghao and Darr, Justin and Zhang, Michael},
title = {Predicting Traffic Performance during a Wildfire Using Machine Learning},
journal = {Transportation Research Record: Journal of the Transportation Research Board},
year = {2023},
volume = {2677},
number = {3},
pages = {1625--1636},
doi = {10.1177/03611981221126509}
}

@article{RahmanHasan2023,
author = {Rahman, Rezaur and Hasan, Samiul},
title = {A Deep Learning Approach for Network-Wide Dynamic Traffic Prediction during Hurricane Evacuation},
journal = {Transportation Research Part C: Emerging Technologies},
year = {2023},
volume = {152},
pages = {104126},
doi = {10.1016/j.trc.2023.104126}
}

@article{ZhangEtAl2024,
author = {Zhang, Xiaojian and Zhao, Xilei and Xu, Yiming and Nilsson, Daniel and Lovreglio, Ruggiero},
title = {Situational-Aware Multi-Graph Convolutional Recurrent Network ({SA-MGCRN}) for Travel Demand Forecasting during Wildfires},
journal = {Transportation Research Part A: Policy and Practice},
year = {2024},
volume = {190},
pages = {104242},
doi = {10.1016/j.tra.2024.104242}
}

@article{RashidEtAl2025,
author = {Rashid, Md Mobasshir and Tirtha, Sudipta Dey and Eluru, Naveen and Hasan, Samiul},
title = {Understanding Hurricane Evacuation Behavior from {Facebook} Data},
journal = {International Journal of Disaster Risk Reduction},
year = {2025},
volume = {116},
pages = {105147},
doi = {10.1016/j.ijdrr.2024.105147}
}

@article{MaLee2025,
author = {Ma, Fangjiao and Lee, Ji Yun},
title = {Analyzing Wildfire Evacuation Dynamics with Agent-Based Modeling in Damaged Road Networks},
journal = {Safety Science},
year = {2025},
volume = {187},
pages = {106835},
doi = {10.1016/j.ssci.2025.106835}
}

@article{JanfeshanaraghiEtAl2025,
author = {Janfeshanaraghi, Nima and Rohaert, Arthur and Ronchi, Enrico and B{'e}nichou, Noureddine and Kuligowski, Erica D.},
title = {Traffic Performance Indicators for Evacuation: The Case Study of the 2020 {Silverado} Wildfire},
journal = {Fire Technology},
year = {2025},
volume = {61},
pages = {5815--5839},
doi = {10.1007/s10694-025-01813-y}
}

@article{JiangEtAl2026,
author = {Jiang, Shangkun and Lovreglio, Ruggiero and Cova, Thomas J. and Park, Sangung and Xu, Susu and Zhao, Xilei},
title = {Wildfire Evacuation Analysis Using {Facebook} Data: Evidence from {Palisades} and {Eaton} Fires},
journal = {Transportation Research Part D: Transport and Environment},
year = {2026},
volume = {158},
pages = {105456},
doi = {10.1016/j.trd.2026.105456}
}

@inproceedings{LiEtAl2018DCRNN,
author = {Li, Yaguang and Yu, Rose and Shahabi, Cyrus and Liu, Yan},
title = {Diffusion Convolutional Recurrent Neural Network: Data-Driven Traffic Forecasting},
booktitle = {Proceedings of the International Conference on Learning Representations},
year = {2018},
url = {https://openreview.net/forum?id=SJiHXGWAZ}
}

@inproceedings{YuEtAl2018STGCN,
author = {Yu, Bing and Yin, Haoteng and Zhu, Zhanxing},
title = {Spatio-Temporal Graph Convolutional Networks: A Deep Learning Framework for Traffic Forecasting},
booktitle = {Proceedings of the Twenty-Seventh International Joint Conference on Artificial Intelligence},
year = {2018},
pages = {3634--3640},
doi = {10.24963/ijcai.2018/505}
}

@misc{CALFIREPalisades2025,
author = {{California Department of Forestry and Fire Protection}},
title = {Palisades Fire},
year = {2026},
howpublished = {\url{https://www.fire.ca.gov/incidents/2025/1/7/palisades-fire}},
note = {Last updated April 9, 2026; accessed July 27, 2026}
}

@misc{CALFIREEaton2025,
author = {{California Department of Forestry and Fire Protection}},
title = {Eaton Fire},
year = {2026},
howpublished = {\url{https://www.fire.ca.gov/incidents/2025/1/7/eaton-fire}},
note = {Last updated June 15, 2026; accessed July 27, 2026}
}

@inproceedings{ShiEtAl2021,
author = {Shi, Yunsheng and Huang, Zhengjie and Feng, Shikun and Zhong, Hui and Wang, Wenjin and Sun, Yu},
title = {Masked Label Prediction: Unified Message Passing Model for Semi-Supervised Classification},
booktitle = {Proceedings of the Thirtieth International Joint Conference on Artificial Intelligence},
year = {2021},
pages = {1548--1554},
doi = {10.24963/ijcai.2021/214}
}

@article{HochreiterSchmidhuber1997,
author = {Hochreiter, Sepp and Schmidhuber, J{"u}rgen},
title = {Long Short-Term Memory},
journal = {Neural Computation},
year = {1997},
volume = {9},
number = {8},
pages = {1735--1780},
doi = {10.1162/neco.1997.9.8.1735}
}

@article{pedregosa2011scikit,
  title={Scikit-learn: Machine learning in Python},
  author={Pedregosa, Fabian and Varoquaux, Ga{\"e}l and Gramfort, Alexandre and Michel, Vincent and Thirion, Bertrand and Grisel, Olivier and Blondel, Mathieu and Prettenhofer, Peter and Weiss, Ron and Dubourg, Vincent and Vanderplas, Jake and Passos, Alexandre and Cournapeau, David and Brucher, Mathieu and Perrot, Matthieu and Duchesnay, {\'E}douard},
  journal={Journal of Machine Learning Research},
  volume={12},
  pages={2825--2830},
  year={2011}
}

@article{rashid2024evacuation,
  author    = {Md. Mobasshir Rashid and Rezaur Rahman and Samiul Hasan},
  title     = {Network-Wide Evacuation Traffic Prediction in a Rapidly Intensifying Hurricane from Traffic Detectors and Facebook Movement Data: Deep-Learning Approach},
  journal   = {Journal of Transportation Engineering, Part A: Systems},
  volume    = {151},
  number    = {1},
  year      = {2024},
  doi       = {10.1061/JTEPBS.TEENG-8416},
  url       = {https://doi.org/10.1061/JTEPBS.TEENG-8416}
}

@misc{caltrans_pems,
  author       = {{California Department of Transportation}},
  year         = {2026},
  title        = {{Caltrans Performance Measurement System (PeMS)}},
  howpublished = {\url{https://pems.dot.ca.gov/}},
  note         = {Accessed July 30, 2026}
}

@article{McConnell2026,
  author  = {McConnell, Kathryn and Mueller, J. Tom and Burow, Paul Berne and St. Denis, Lise Ann},
  title   = {Urban Areas in the United States Experience Substantial Wildfire Impacts},
  journal = {Communications Earth \& Environment},
  year    = {2026},
  volume  = {7},
  number  = {1},
  pages   = {352},
  doi     = {10.1038/s43247-026-03336-y},
  url     = {https://doi.org/10.1038/s43247-026-03336-y},
  issn    = {2662-4435}
}

@inproceedings{ZhuEtAl2020,
  author    = {Zhu, Jiong and Yan, Yujun and Zhao, Lingxiao
               and Heimann, Mark and Akoglu, Leman and Koutra, Danai},
  title     = {Beyond Homophily in Graph Neural Networks:
               Current Limitations and Effective Designs},
  booktitle = {Advances in Neural Information Processing Systems},
  volume    = {33},
  pages     = {7793--7804},
  year      = {2020},
  url       = {https://proceedings.neurips.cc/paper/2020/hash/58ae23d878a47004366189884c2f8440-Abstract.html}
}

\end{document}